\documentclass[runningheads]{llncs}

\usepackage{eccv}

\usepackage{eccvabbrv}

\usepackage{graphicx}
\usepackage{booktabs}
\usepackage{placeins}
\usepackage[accsupp]{axessibility}  

\usepackage{hyperref}

\usepackage{orcidlink}

\usepackage{multirow}
\usepackage{graphicx}
\usepackage{paralist}

\usepackage{chngcntr}

\def\name{GENA3D} 

\usepackage{tcolorbox}
\tcbuselibrary{skins, breakable}
\usepackage{xcolor}

\definecolor{systembg}{HTML}{F0F4FF}
\definecolor{userbg}{HTML}{F6FFF6}

\usepackage{listings}
\newtcolorbox{promptbox}{
  breakable,
  colback=gray!10,
  colframe=gray!50,
  boxrule=0.5pt,
  fontupper=\small\ttfamily,
  before skip=10pt,
  after skip=10pt,
}

\newcommand{\junwei}[1]{\textcolor{black}{#1}} 

\newcommand{\purpleurl}[1]{%
    \href{#1}{\textcolor{purple}{#1}}
}

\begin{document}

\title{\name{}: Generative Amodal 3D Modeling by  \\ Bridging 2D Priors and 3D Coherence} 

\titlerunning{GENA3D}

\author{Junwei Zhou\orcidlink{0009-0008-7093-2280} \and
Yu-Wing Tai\orcidlink{0000-0002-3148-0380}}

\authorrunning{J. Zhou and Y.-W. Tai}

\institute{Dartmouth College \\ \vspace{1em}\textit{\purpleurl{https://colezwhy.github.io/gena3d/}}}

\maketitle

\begin{figure}[t]
    \centering
    \small
    \includegraphics[width=1\linewidth]{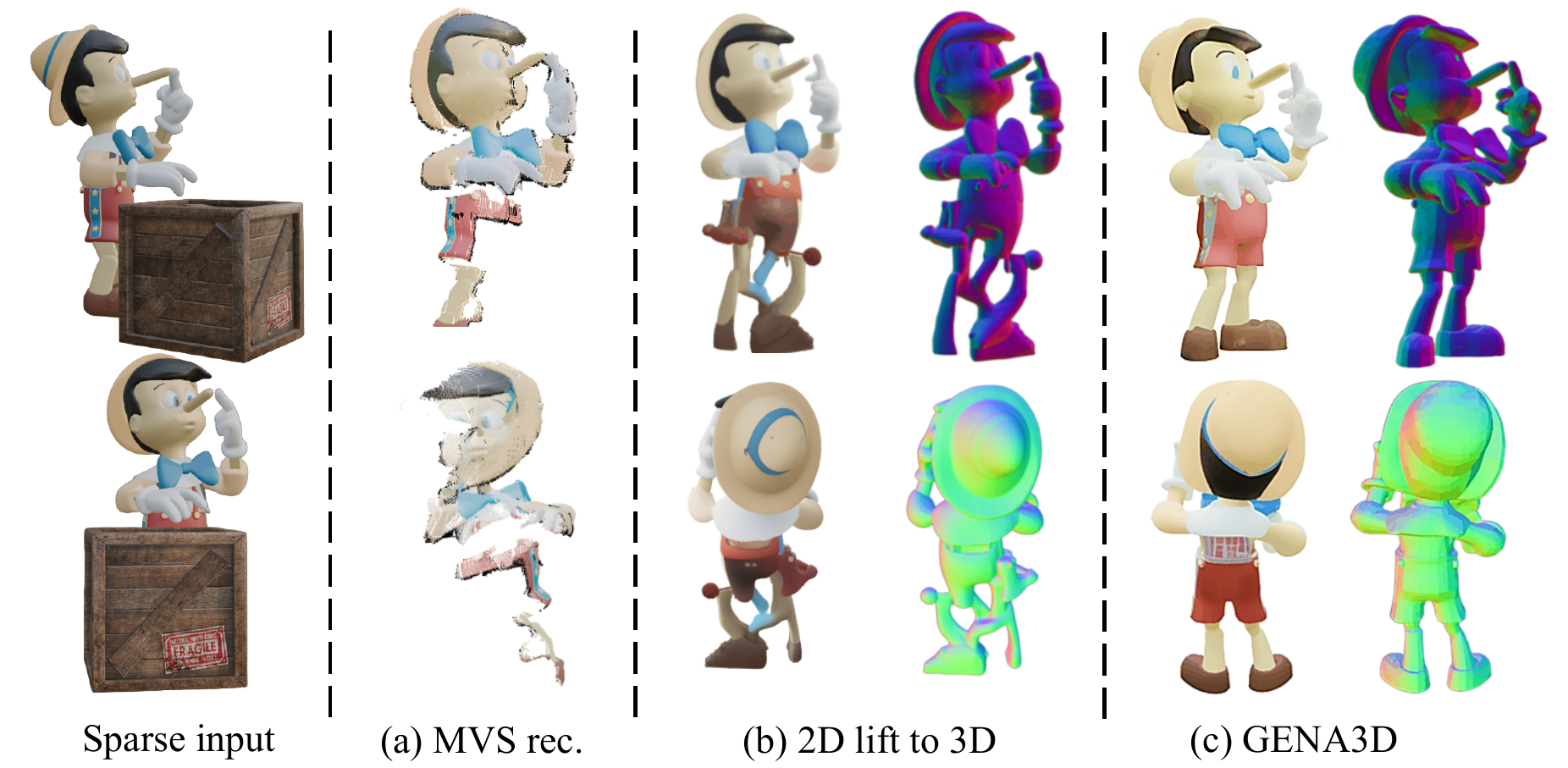}\\
    \vspace{-0.1in}
    \caption{
    Comparison of 3D object reconstruction/generation from sparse and partial occluded inputs.
    From left to right: input views, results from (a) MVS-based reconstruction (\textit{e.g.}, VGGT~\cite{wang2025vggt}), (b) 2D amodal completion lifted to 3D (TRELLIS~\cite{xiang2025structured}), and (c) \textbf{\name{} (ours)}.
    Our model faithfully generate complete, occlusion-free 3D geometry consistent with input partially visible constraints.
    }\vspace{-0.15in}
    \label{fig:teaser}
\end{figure}

\vspace{-.1in}
\begin{abstract}
Generating complete 3D objects under partial occlusions (\ie, amodal scenarios) is a practically important yet challenging problem, as large portions of object geometry are unobserved in real-world scenarios.
Existing approaches either operate directly in 3D, which ensures geometric consistency but often lacks generative expressiveness, or rely on 2D amodal completion, which provides strong appearance priors but does not guarantee reliable 3D structure. This raises a key question: \textit{how can we achieve both generative plausibility and geometric coherence in amodal 3D modeling?}
To answer this question, we introduce \name{} (\textbf{GEN}arative \textbf{A}modal 3D), a framework that integrates learned 2D generative priors with explicit 3D geometric reasoning within a conditional 3D generation paradigm. The 2D priors enable the model to plausibly infer diverse occluded content, while the 3D representation enforces multi-view consistency and spatial validity. Our design incorporates a novel View-Wise Cross-Attention for multi-view alignment and a Stereo-Conditioned Cross-Attention to anchor generative predictions in 3D relationships.
By combining generative imagination with structural constraints, \name{} generates complete and coherent 3D objects from limited observations without sacrificing geometric fidelity. Experiments demonstrate that our method outperforms existing approaches in both synthetic and real-world amodal scenarios, highlighting the effectiveness of \junwei{bridging 2D priors and 3D coherence} in generating plausible and geometrically consistent 3D structures in complex environments.

  \keywords{\junwei{Amodal 3D Object Generation} \and \junwei{Generative Models} }
\end{abstract}

\section{Introduction}
\label{sec:intro}

\junwei{Generating complete 3D objects under partial occlusions} is a fundamental but challenging task in computer vision. In \junwei{real-world} scenarios, objects are often captured from arbitrary viewpoints, with \junwei{certain} portions hidden by surrounding structures or other objects. Such incomplete observations make the \junwei{problem of generating 3D objects in the scene} inherently amodal: a system must infer the geometry and appearance of regions that are never directly observed \junwei{in the input images}. This ability is crucial for object-level 3D scene understanding and downstream applications \junwei{that are critically dependent on such capability,} such as robotic interaction, AR/VR content creation, and embodied AI perception.

\junwei{Recent advances in 3D object generation have been driven by diffusion-based and autoregressive models that learn powerful structural and semantic priors from large-scale 3D datasets. These approaches can synthesize diverse and realistic shapes from text prompts, single images, or even sparse inputs. However, most existing 3D generative models heavily depend on learned strong geometric priors and implicitly assume high-quality observations and canonical object poses to produce high-quality results.
When confronted with sparse and partially occluded real-world inputs, these models often struggle to effectively leverage the limited visible evidence to plausibly infer the missing regions. The inherent complexity of the 3D domain and the scarcity of corresponding training data further exacerbate this issue, leading to inconsistent geometry, implausible structures, loss of fine-grained details, or even catastrophic generation failures.}

To address this challenge, we introduce \textbf{\name{}}, a generative framework for amodal 3D object generation from sparse, unposed views. 
Our approach treats this problem as a conditional generation process that \junwei{bridges 2D priors and 3D coherence}. 
Specifically, we first perform 2D object-centric amodal completion that are rich in generative ability on each view to recover occluded regions. 
The features of completed but inconsistent views are then aggregated through a \emph{View-Wise Cross Attention} module that produces model latents fused from sparse-view information. 
%
To ensure the fidelity of visible regions and 3D coherence, we condition the generation on partial geometry (point clouds) estimated from an MVS model. 
This information is incorporated through a \emph{Stereo-Conditioned Cross Attention} module equipped with a geometry-guided gating mechanism, which modulates fused latent based on 3D spatial cues. 
After processing the generated sparse structure through another pretrained Flow Transformer, the resulting structured latent representation is then decoded to produce a complete, occlusion-free 3D model that \junwei{faithfully} respects to the \junwei{observed geometry} while plausibly \junwei{generating} the hidden regions. 
As shown in Fig.~\ref{fig:teaser} (c), our method produces coherent and complete \junwei{3D objects} even when large \junwei{portions} of the \junwei{geometry} are never observed or the 2D completions are not consistent.

Extensive experiments on both synthetic and real-world datasets demonstrate that \textbf{\name{}} achieves superior \junwei{generation} quality compared with existing sparse-view and amodal \junwei{3D generation} baselines. 
Our method improves geometric \junwei{faithfulness} while maintaining realistic appearance, achieving consistent \junwei{and superior} performance from single-view to multi-view inputs. 
These results highlight the potential of generative modeling for practical 3D object-level scene \junwei{synthesis} under real-world occlusion and view sparsity.

The contribution of this work is threefold:
\begin{compactitem}
    \item We introduce \textbf{\name{}}, a \junwei{generative amodal 3D modeling framework that} recover complete geometry and appearance from sparse and occluded inputs by bridging 2D priors and 3D coherence during generation.
    \item \textcolor{black}{We propose two principled conditioning mechanisms for amodal 3D generation: (1) \emph{View-Wise Cross Attention}, which introduces multi-view reasoning to mitigate view dominance and geometric drift; and (2) \emph{Stereo-Conditioned Cross Attention}, which integrates partial stereo geometry through logit-level gating, transforming geometric cues into explicit attention regulators for structurally coherent generation.}
    \item Comprehensive experiments validate that \name{} achieves superior \junwei{generation} fidelity, completeness, and consistency compared with existing sparse-view and amodal \junwei{generation} methods.
\end{compactitem}

\section{Related Work}
\label{sec:related}

\noindent\textbf{3D Generative Models.}
Recent advances in 3D generative models have opened new possibilities for structure-aware reconstruction.
Methods leveraging Gaussian Splatting~\cite{kerbl20233d} and NeRF~\cite{mildenhall2021nerf} as these representations excel at novel view synthesis~\cite{ye2024no, truong2023sparf}, but struggle to \junwei{segregate object-level entities}, hindering object-level 3D scene understanding.
Diffusion-based generative models
further enhance visual fidelity, controllability, and scalability by leveraging pretrained 2D diffusion priors\junwei{~\cite{huang2024dreamcontrol, ding2024bidirectional, chen2024vp3d}} or 3D-aware latent representations\junwei{~\cite{xiang2025structured, long2024wonder3d, zhao2025hunyuan3d, lin2025kiss3dgen, siddiqui2024assetgen}}. These methods typically rely on \junwei{massive optimization iterations or decently posed input images for proper results.}
%
Some works attempt to mitigate missing regions using inpainting-augmented generation~\cite{chen2024comboverse, han2024reparo, zhou2024layout, zhou2025coco4dcomprehensivecomplex4d}.
Other scene-level generative pipelines~\cite{chen2024single, liu2022towards, huang2025midi, meng2025scenegen, yao2025cast} synthesize entire 3D scenes from a single image, but they rely on carefully curated data and pose-calibrated views, \junwei{which limits their effectiveness under real-world, unposed scenarios.}

{In contrast to prior approaches, \name{} is a generative 3D reconstruction model tailored for sparse, unposed inputs. By fusing generative priors with partial geometric conditioning, it produces reconstructions that are both geometrically faithful and semantically coherent, remaining consistent with sparse-view constraints while plausibly inferring unseen object regions.
}

\vspace{2mm}
\noindent\textbf{Amodal Completion in 2D/3D.}
Amodal completion aims to infer the hidden or occluded parts of objects beyond visible evidence. Early works address this through explicit amodal segmentation and completion tasks~\cite{LiM_2016_ECCV, Qi_2019_CVPR, Ling_2020_NeurIPS, Mohan_2022_CVPR, Li_2023_ICCV_MUVA}, focusing primarily on 2D perception and annotation-based benchmarks. 
These methods rely on supervised mask prediction or latent shape priors to hallucinate occluded object regions, but lack in fidelity and geometric consistency.

Generative 2D inpainting models~\cite{Rombach_2022_CVPR, yu2023inpaint} have since \junwei{become common solutions} in image-level amodal completion, synthesizing realistic missing content based on surrounding context. 
Nowadays, researchers further enhance this capacity by introducing specialized amodal reasoning modules~\cite{ao2025open, lee2025tuningfreeamodalsegmentationocclusionfree, fan2025multi} or by fine-tuning diffusion models for occlusion-aware reconstruction~\cite{ozguroglu2024pix2gestalt, xu2024amodal, lu2025taco}. Extending beyond single-view imagery, multi-view inpainting models~\cite{barda2025instant3dit, weber2023nerfiller} attempt to maintain cross-view coherence but often assume calibrated cameras or dense view coverage. 
More geometry-aware approaches~\cite{Li_2022_ECCV, salimi2025geometry, feng2025objfiller, shi2025imfine, Zhan_2024_CVPR} leverage 3D representation priors for scene-level completion; however, they remain limited in handling object-level amodal generation due to alignment errors and weak semantic reasoning. 
The most relevant work is Amodal3R~\cite{wu2025amodal3r}, which trains a native 3D \junwei{generation model} from \junwei{occluded} objects, but it is constrained by data scale and limited occlusion reasoning ability.

In sparse and unposed scenarios, 2D inpainting models can recover locally plausible structures but often introduce cross-view inconsistencies that lead to geometric artifacts in 3D reconstruction. Our method integrates strong 2D amodal priors into a 3D generative framework, enforcing global geometric coherence via view-wise and stereo-conditioned attention. Compared with single-image 3D amodal methods, \name{} enables more robust reconstruction under severe occlusion by aggregating multi-view information and enhancing cross-view geometric reasoning, while preserving realistic image-level completion.

%

\noindent\textbf{Multi-View Stereo (MVS).}
Multi-view stereo (MVS)~\cite{FurukawaHernandez2015_MultiViewStereoTutorial, Wang2024_LearningBasedMVS_Survey, Yi2020PyramidMVSNet, Yan2020DenseHybridRMVSNet, wang2024dust3r, leroy2024grounding, zhang2024monst3r, yang2025fast3r, zhang2025flare, wang2025pi, keetha2025mapanything} has been studied for decades as a core technique for 3D reconstruction from multiple \junwei{input views}. 
\junwei{Recent advances~\cite{Ren_2023_HPMMVS, Wu_2024_GoMVS, Santo_2024_MVCPS_NeuS}} in \emph{dense correspondence learning} have enhanced MVS accuracy and robustness by jointly reasoning about geometry and radiance. 
%
%
These methods achieve fine-grained alignment but remain limited \junwei{when it comes to} calibrated settings and fixed depth ranges. 
%
\junwei{To improve scalability and avoid per-scene optimization, 
\emph{feed-forward extensions}~\cite{Izquierdo_2025_MVSAnywhere, Cabon_2025_MUSt3R, Liu_2024_MVSGaussian, Dong_2024_DepthRangeFreeMVS} introduce transformer-based architectures and specific techniques for zero-shot multi-view reconstruction.} 
These designs collectively enhance scalability and inference efficiency. 
%


\textcolor{black}{Despite substantial progress, MVS methods are primarily designed for \emph{reconstruction} rather than \emph{generation}. In contrast, we do not seek to improve MVS. Instead, we leverage partial MVS geometry as an explicit structural prior within a generative 3D framework. By conditioning on stereo-derived point clouds through our Stereo-Conditioned Cross Attention, \name{} converts incomplete geometric cues into attention regulators that guide plausible and geometrically coherent completion. This effectively bridges feed-forward MVS geometry with conditional 3D generation for object-level amodal modeling.}

%

\begin{figure*}[!t]
    \centering
    \includegraphics[width=1.0\linewidth]{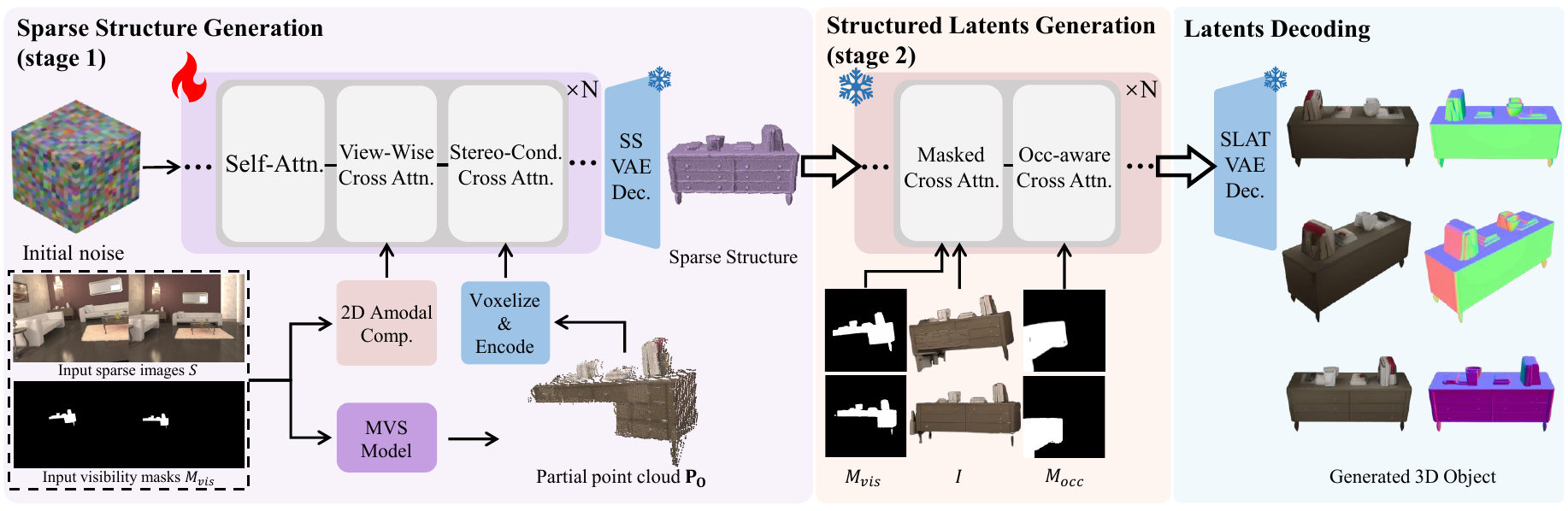}\\
    \vspace{-.05in}
    \caption{\textbf{Overview of \name}. Given sparse images $S$, visibility masks $M_{vis}$, and occlusion masks $M_{occ}$ indicating the occluded object $O$, \name{} first generates a sparse structure (stage 1) by aggregating multi-view information in Sec.~\ref{method:viewwise} and infers the complete geometric structure from the partial stereo point cloud $\mathbf{P}_{O}$ in Sec.~\ref{method:scca}. We then employ a pretrained amodal SLAT Transformer to generate details with $M_{vis}$ and $M_{occ}$, and then generate the structured latent. This structured latent is decoded into an occlusion-free 3D object with high-quality geometry and appearance.}
    \label{fig:model}
    \vspace{-.05in}
\end{figure*}

\section{Method}
\label{sec:method}
For an arbitrary object $O$ in the given $K$ sparse views $S = \{s^1, s^2, ..., s^K\}$ (typically 1-4 views), we can obtain the visibility masks of $O$ and the corresponding text prompt automatically or manually using vision foundation models, \eg, SAM \cite{kirillov2023segment} and Florence 2 \cite{xiao2024florence}. 
After that, we follow the \junwei{occlusion analysis and 2D amodal completion strategies from} \cite{ao2025open} with minor modifications (using the above obtained visibility mask and prompt) to first perform generative image-level amodal completion across all views using ~\cite{ao2025open} \junwei{(denoted as OAAC)}.
Altogether, we can obtain visual clues of the desired object $O$ at all $K$ input views: the visibility masks $M_{vis}=\{m^1_{v}, m^2_{v}, ..., m^K_{v}\}$ indicating the visible part of $O$; the inconsistent 2D completions of $O$: $I=\{i^1, i^2, ..., i^K\}$; and the occlusion masks $M_{occ} = \{m^1_{o}, m^2_{o}, ..., m^K_{o} \}$ \junwei{representing} the potential occlusion regions.

We illustrate the overall model structure in Fig.~\ref{fig:model}. 
Our design prioritizes the Sparse Structure Transformer, \junwei{\ie, stage 1}. Given object-centric sparse views $I$ \junwei{processed using 2D amodal completion methods}, we apply the proposed \textbf{View-Wise Cross Attention} (Sec.~\ref{method:viewwise}) to aggregate multi-view features and fuse their latents. The fused representation is further refined through the \textbf{Stereo-Conditioned Cross Attention} (Sec.~\ref{method:scca}) for unobserved structure inference, guided by \junwei{observed} partial stereo geometry and a geometry-aware gating mechanism. The resulting Sparse Structure is \junwei{decoded and} then passed to the pretrained Structure Latent Transformer from \cite{wu2025amodal3r} for Structured Latent generation. Implementation details and training data preparation 
are described in Sec.~\ref{method:training}.



\begin{figure}[t]
\begin{minipage}[c]{0.32\linewidth}
    \includegraphics[width=\linewidth]{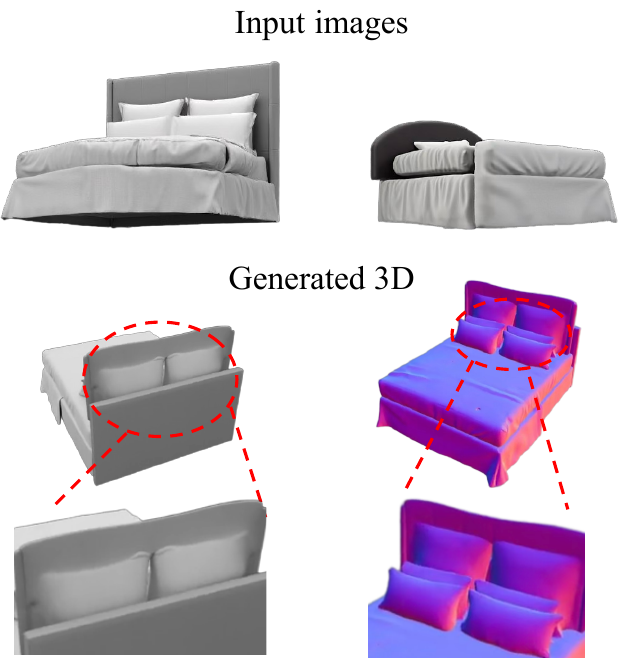}\\
    \vspace{-0.15in}
        \caption{\small The original model shows a strong bias and results in artifact accumulation when conditioned with multi-view.}
    \label{fig:trellis_multi-view}
\end{minipage}
\hfill
\begin{minipage}[c]{0.65\linewidth}
    \includegraphics[width=\linewidth]{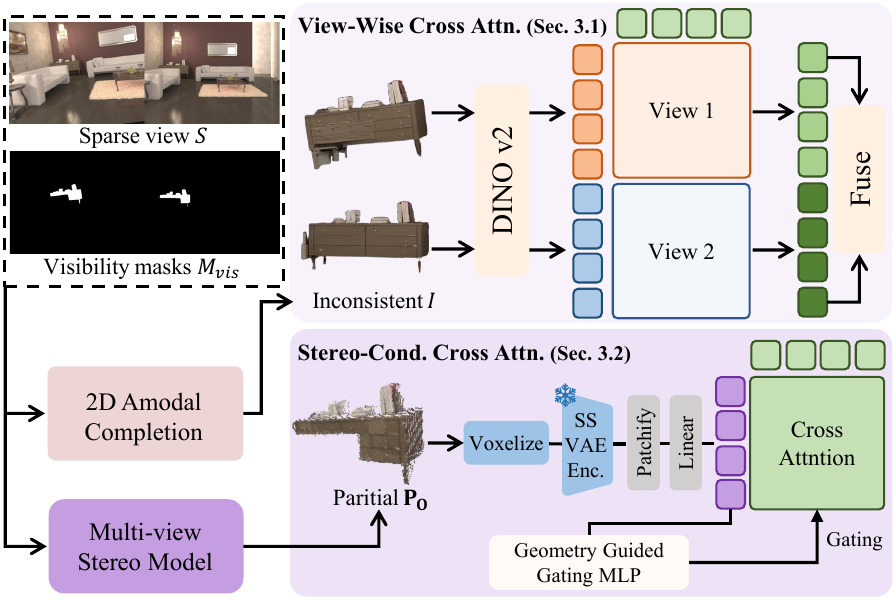}\\
    \vspace{-0.25in}
  \caption{\small A detailed illustration of our proposed View-Wise Cross Attention and Stereo-Conditioned Cross Attention modules.}
  \label{fig:pipeline_proposed}
\end{minipage}
\vspace{-.1in}
\end{figure}

\subsection{View-Wise Cross Attention}
\vspace{-.1in}
\label{method:viewwise}
\textcolor{black}{Existing multi-view conditioning strategies in generative models typically adopt either sequential cross-attention (conditioning on one view per sampling step) or naïve feature concatenation across views. As shown in Fig.~\ref{fig:trellis_multi-view}, under sparse and unposed amodal settings, these strategies introduce two critical issues: (1) view dominance, where later-conditioned views overwrite earlier structural hypotheses, and (2) geometric drift, where inconsistencies in independently completed 2D views accumulate into unstable 3D structure.} 


\textcolor{black}{To address this, we propose View-Wise Cross Attention, a parallelized multi-view conditioning mechanism that enforces simultaneous and balanced cross-view reasoning at every generation step.}
\junwei{Since the base 3D generation model consistently produces objects (latents) in canonical poses regardless of the conditions' camera poses, we exploit this pose-invariant property as a prior in designing our multi-view fusion strategy.}
We denote the intermediate latent as $z \in \mathbb{R}^{L \times D}$, where $L$ is the number of tokens and $D$ is the hidden dimension \junwei{of the tokens}.
In the standard Cross Attention (denoted as $\mathcal{CA}$), the latent $z$ interacts with the encoded image feature $c$,
\begin{align}
c = \texttt{DINO}(i), \quad z' = \mathcal{CA}(z, c).
\label{eq1: origin_ca}
\end{align}
Here, DINOv2~\cite{oquab2023dinov2} serves as the image encoder, \junwei{extracting features for input image $i$}. When multiple images are available,
%
for each 2D completion view $i^n$, we extract its DINO feature $c^n$, forming a set of encoded features $C = [c^1, c^2, ..., c^K]$. We then apply Cross Attention in parallel for each view:
\begin{align}
z_n' = \mathcal{CA}(z, c^n), \quad \forall n \in {1, 2, \dots, K}.
\label{eq2:eachview}
\end{align}
The resulting view-wise latents $Z' = [z_1', z_2', ..., z_K']$ are fused into a single latent $z'$ through a visibility-weighted average:
\begin{align}
\!\!\!\!z' = \frac{1}{K} \sum_{n=1}^{K} w_n z_n', \;
w_n = \frac{\tau_n}{\sum_{j=1}^{K}\tau_j}, \;
\tau_n = \frac{m_v^n}{m_o^n + m_v^n}.
\label{eq3:meanavg}
\end{align}
$\tau_n$ represents the visibility ratio, defined as the proportion of possibly visible region in view $i^n$. A final LayerNorm is applied after fusion for numerical stability.

\textcolor{black}{By structurally decoupling per-view reasoning and explicitly rebalancing contributions using geometric visibility priors, View-Wise Cross Attention converts inconsistent 2D amodal completions into a coherent shared latent space, significantly improving cross-view stability under sparse observations.}



\subsection{Stereo-Conditioned Cross Attention Module}
\label{method:scca}
\textcolor{black}{While prior generative models may incorporate geometric signals (e.g., depth maps or point clouds) as additional tokens, they typically treat geometry as passive conditioning features. This often results in weak structural enforcement, particularly when the geometry is partial, noisy, or unposed.
To address this, we introduce Stereo-Conditioned Cross Attention, a geometry-aware attention mechanism that actively modulates the attention distribution using encoded stereo structure. Instead of merely concatenating geometric tokens, Stereo-Conditioned Cross Attention injects geometric priors directly into the attention logits via a geometry-guided gating function.}

Given sparse observations $S=\{s^1, s^2, ..., s^K\}$, we obtain the stereo point cloud $\mathbf{P}=\{p^1, p^2, ..., p^K\}=\texttt{MVS}(S)$, where each $p^n$ corresponds to a point cloud reconstructed from view $s^n$, \junwei{by lifting each pixel to the 3D space using the depth value}. We extract the partial point cloud of the target object by selecting points projected within the visibility masks:
\begin{align}
    \mathbf{P}_O = \bigcup_{n=1}^{K} p_v^n, \quad p_v^n = \{\, p \in p^n \mid m_v^n(p)=1 \}.
    \label{eq5:visible_points}
\end{align}
The resulting $\mathbf{P}_O$ is voxelized into active voxels $\mathbf{V}_O$ with grid size $\varphi$, then encoded by the Sparse Structure VAE, and finally transformed into geometric feature tokens \junwei{using the same patchify and linear layer for input transformation}:
\begin{equation}
\begin{aligned}
    \mathbf{V}_O &= \texttt{Voxelize}(\mathbf{P}_O, \varphi), \quad
    f_{\mathbf{V}_O} = \mathbf{Enc}(\mathbf{V}_O), \\
    c_{\text{geo}} &= \texttt{Linear}(\texttt{Patchify}(f_{\mathbf{V}_O})).
\end{aligned}
\label{eq6:geo_condition}
\end{equation}
We further introduce a \textbf{Geometry-Guided Gating MLP} that modulates attention using the encoded geometry:
\begin{equation}
\begin{aligned}
g &= \sigma(\text{MLP}_{gating}(c_{\text{geo}})), \\
\alpha &= \texttt{softmax}\!\left(\frac{QK^\top}{\sqrt{D}} + \log(g+\varepsilon)\right),
\label{eq7:gating_mlp}
\end{aligned}
\end{equation}
where $\alpha$ denotes the attention weights computed from query $Q=W_q z'$ and key $K=W_k c_{\text{geo}}$, and $\varepsilon$ is a small constant for logarithmic numerical stability.

\textcolor{black}{By embedding 3D structure directly into the attention computation, Stereo-Conditioned Cross Attention establishes a tighter coupling between generative imagination and geometric plausibility, enforcing structural coherence even when large object regions remain unobserved.}


\subsection{Data Engine \& Training/Inference Proxy}
\label{method:training}
\noindent\textbf{Data Engine.} We leverage object-centric 3D datasets from~\cite{xiang2025structured} and convert them into watertight meshes using Manifold~\cite{huang2018robust}. 
To simulate realistic occlusions, we generate 3D-consistent occlusion masks by randomly selecting a seed face on each mesh and iteratively expanding to its neighboring faces. 
This process continues outward until reaching a randomly sampled coverage ratio, producing spatially coherent and natural occlusion patterns. 
We then render object views and their corresponding occlusion masks, discarding camera poses with extreme angles or insufficient occlusion. 
Finally, completed object views are synthesized by treating the occlusion masks as inpainting regions and applying controlled mask distortions for greater diversity (see supplementary for more details).

\noindent\textbf{Training Proxy.} During training, the number of sparse views per object is randomly sampled from $\{1,2,3,4\}$ for each batch. 
After sampling, stereo point clouds are generated online based on the selected views. 
The resulting partial point clouds are centralized and normalized to ensure training stability. 
Thanks to the synthesized occluded object inputs and visibility masks, modern multi-view stereo models can still recover plausible partial geometry, providing reliable geometric supervision throughout training. 

We optimize the model using the standard conditional flow matching (CFM)~\cite{lipman2023flowmatchinggenerativemodeling} objective during training:
\begin{equation}
\begin{aligned}
    \mathcal{L}_{\text{CFM}}(\theta) = 
    \mathbb{E}_{t, z_0, \epsilon} 
     \| v_\theta(z, t) - (\epsilon - z_0) \|_2^2 ,
    \label{eq8:flow}
\end{aligned}
\end{equation}
where $\theta$ denotes the network parameters, $z_0$ and $\epsilon$ are data and noise samples, $t$ is the timestep, and $z = (1-t)x_0 + t\epsilon$ represents the interpolated state.

\noindent\textbf{Inference Proxy.} During inference, our method can process both scene-level and object-level images. For scene-level images, we first generate the stereo point clouds for the entire scene with $S$, and then extract the partial point clouds for each object using the corresponding visibility masks $M_{vis}$. This approach reduces computational overhead when generating multiple objects within a single scene. All other procedures remain consistent with what is done in the training process. Fig.~\ref{fig:model} shows an example of scene-level inference.

\section{Experiments}
\label{sec:exp}

\begin{figure*}[!t]
    \centering
    \includegraphics[width=1.0\linewidth]{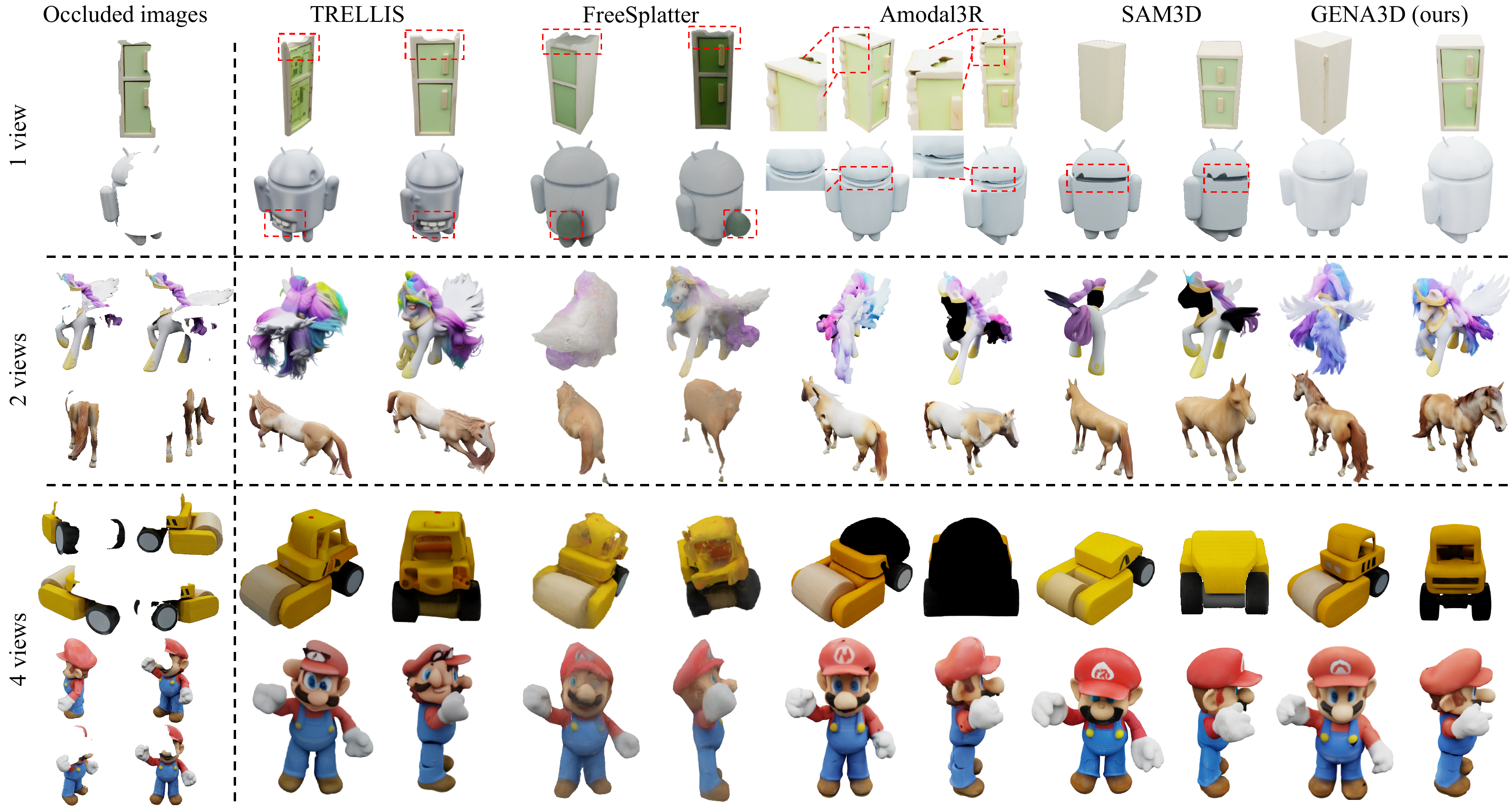}\\
    \vspace{-.05in}
    \caption{Amodal 3D object generation results on the GSO dataset. We give examples generated with 1, 2, and 4 views, respectively. Note that for FreeSplatter~\cite{xu2025freesplatter}, we adopt the Hunyuan-MV ~\cite{yang2024tencent} to generate multiple views under the input of single image.}
    \label{fig:qual_gso}
    \vspace{-.1in}
\end{figure*}

\begin{figure}[t]
\begin{minipage}[c]{0.6\linewidth}
    \includegraphics[width=1.0\linewidth]{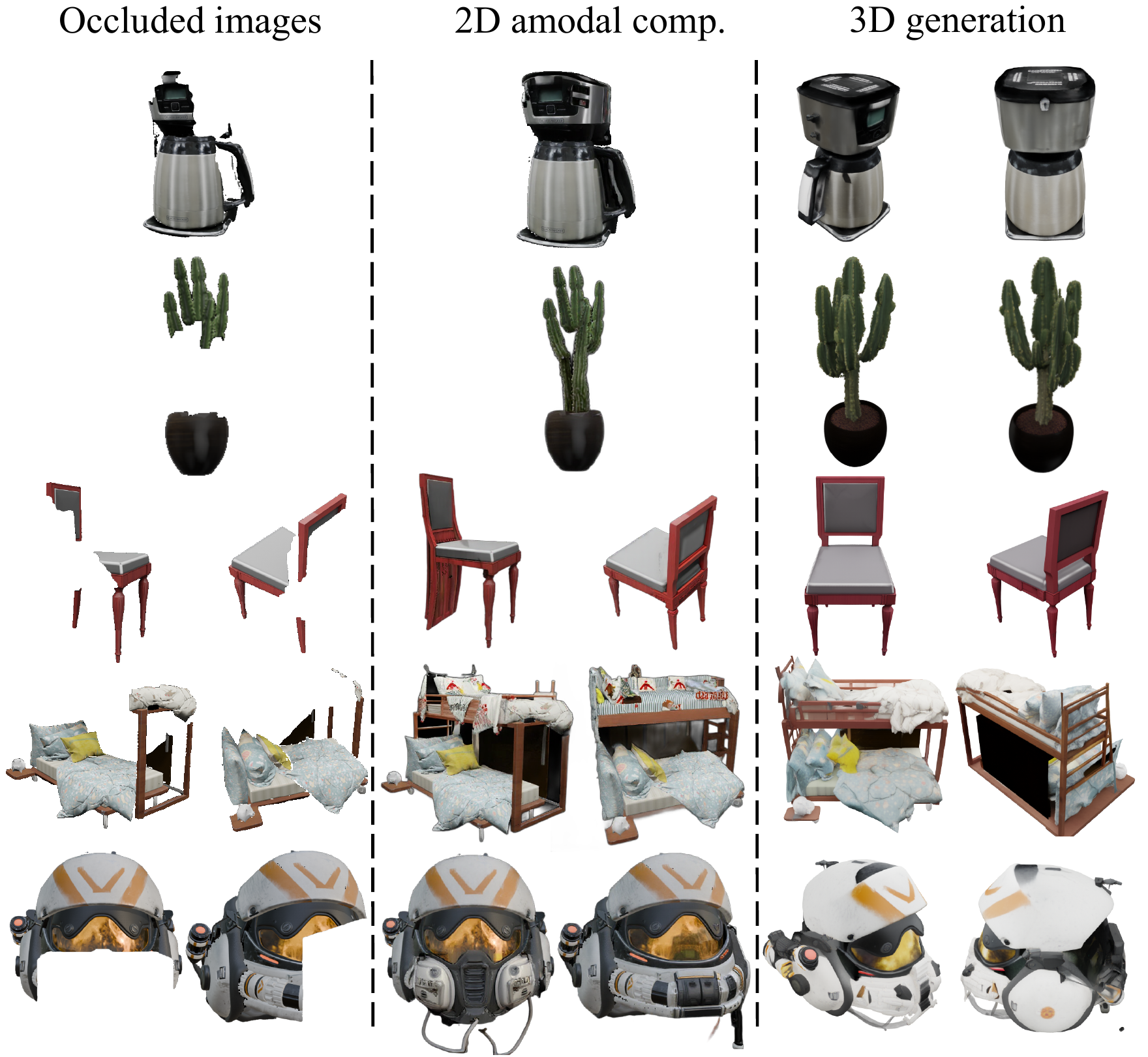}\\
    \vspace{-0.25in}
        \caption{\small Additional qualitative results, together with occluded input images after 2D amodal completion.}
    \label{fig:comp_coarse}
\end{minipage}
\hfill
\begin{minipage}[c]{0.36\linewidth}
    \includegraphics[width=\linewidth]{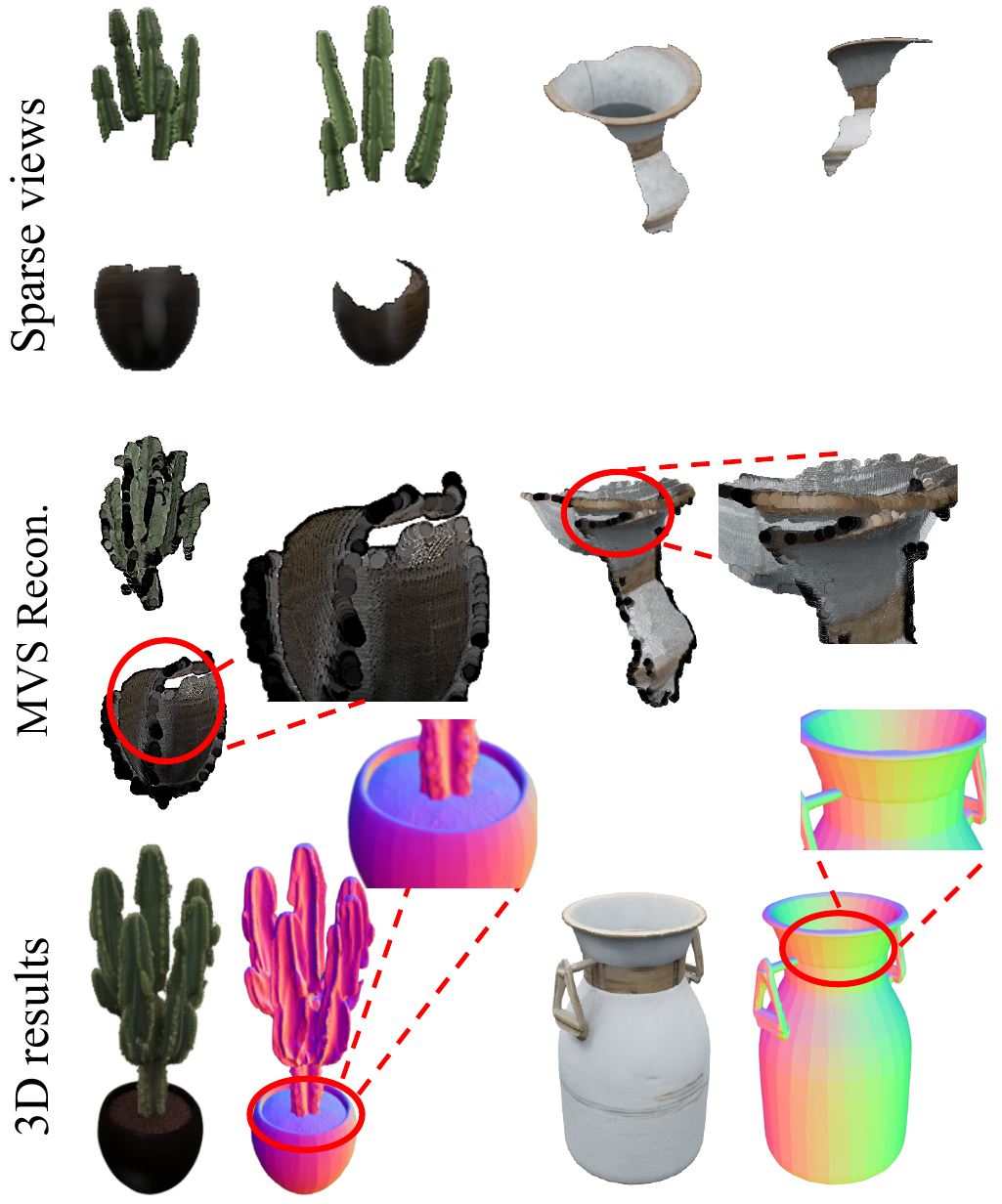}\\
    \vspace{-0.25in}
  \caption{\small We demonstrate that \name{} can resist a certain degree of errors in the MVS reconstructed point clouds, \eg, misalignment and overlapping geometry from different views.}
  \label{fig:ablation_mvs}
\end{minipage}
\vspace{-.05in}
\end{figure}

{\begin{table*}[t]
\centering
\caption{Quantitative results on amodal 3D object generation, validated on the GSO dataset. We compare under the settings of 1, 2, and 4 views. The best and second-best results are 
\textbf{bold} and \underline{underlined}, respectively. For sparse-view settings, we either apply 2D amodal completion to each view using ~\cite{ao2025open} (OAAC), or perform 2D amodal completion on a single view followed by multi-view diffusion generation (OAAC + MV)}
\vspace{-.05in}
\resizebox{\textwidth}{!}{
\begin{tabular}{l|c|c|cccccc}
\toprule
Method & View Num. & 2D Amodal & FID $\downarrow$  & KID{\scriptsize(\%)} $\downarrow$  &CLIP{\scriptsize(\%)} $\uparrow$ & MMD{\scriptsize(\textperthousand)} $\downarrow$  & COV{\scriptsize(\%)} $\uparrow$ & PCS{\scriptsize(\%)}$\uparrow$ \\
\midrule
TRELLIS~\cite{xiang2025structured} & \multirow{5}{*}{1 view} & OAAC & 49.68 & 3.21 & 76.79 & 6.08 & 33.37 & 21.1 \\ 
FreeSplatter\ddag~\cite{xu2025freesplatter} &  & OAAC + MV & 72.88 & 4.68 & 78.14 &  7.24 & 35.78 & 25.5 \\ 
Amodal3R~\cite{wu2025amodal3r} &  & N/A & \underline{39.05} & \textbf{0.44} & \underline{80.06} & \underline{5.52} & 38.64 & 46.3 \\ 
SAM3D~\cite{chen2025sam} & & N/A & 34.68 & \underline{0.46} & 82.16 & \textbf{5.50} & \underline{39.03} & \underline{46.5} \\ 
Ours & & OAAC & \textbf{33.91} & \underline{0.46} & \textbf{82.23} & 5.68 & \textbf{39.19} & \textbf{61.9} \\ 
\midrule
TRELLIS~\cite{xiang2025structured}  &\multirow{4}{*}{2 views} & OAAC & 46.23 & 3.28 & 77.01 & 6.27 & 34.12 & 25.0 \\ 
FreeSplatter~\cite{xu2025freesplatter}  & & OAAC & 91.32 & 6.77 & 74.24 & 10.44 & 27.74 & 12.4\\ 
Amodal3R~\cite{wu2025amodal3r}  & & N/A &  \underline{35.52} & \textbf{0.44} & \underline{81.84} & \underline{5.52} & \underline{38.92} & \underline{47.1}\\
Ours & & OAAC & \textbf{32.12} & \underline{0.45} & \textbf{82.33} & \textbf{5.50} & \textbf{39.08} & \textbf{66.3} \\ 
\midrule
TRELLIS~\cite{xiang2025structured}  &\multirow{5}{*}{4 views} & OAAC & 45.41 & 2.79 & 78.17 & 5.94 & 34.42  & 23.2 \\ 
TRELLIS\ddag~\cite{xiang2025structured} & & OAAC + MV & 44.85 & 1.94 & 76.24 & 6.71 & 34.24 & 28.6 \\
FreeSplatter~\cite{xu2025freesplatter} & & OAAC & 94.10 & 4.64 & 75.58 & 10.48 & 32.31 & 8.4 \\
Amodal3R~\cite{wu2025amodal3r}  &  & N/A & \underline{35.15} & \textbf{0.43} & \underline{82.04} & \underline{5.51} & \underline{38.83} & \underline{46.2} \\ 
Ours & & OAAC & \textbf{30.73} & \textbf{0.43} & \textbf{82.53} & \textbf{5.48} & \textbf{39.48} & \textbf{67.0} \\ 
\bottomrule
\end{tabular}
}
\label{tab: exp_metric}
\vspace{-.2in}
\end{table*}
}

{\begin{table}[ht!]
\centering
\small
\vspace{.05in}
\renewcommand\arraystretch{1}
\caption{The generation faithfulness to the observed visual content (regions within visibility masks $M_{vis}$). We evaluate with 1, 2, and 4 input views.}
\vspace{-.05in}
\begin{tabular}{c|c|@{\hspace{4pt}}c@{\hspace{8pt}}c@{\hspace{8pt}}c}
\toprule
Methods &{\hspace{4pt}} Num. of Views {\hspace{4pt}} & SSIM$\uparrow$  & PSNR$\uparrow$  & LPIPS$\downarrow$ \\
\midrule
 TRELLIS~\cite{xiang2025structured} & \multirow{4}{*}{1 view} & 0.721 & 14.15 & 0.325 \\
 SAM3D~\cite{chen2025sam} &  & 0.797 & 15.54 & 0.226 \\ 
 Amodal3R~\cite{wu2025amodal3r} &  & 0.814 & 16.41 & 0.241 \\ 
  Ours &  & 0.832 & 16.52 & 0.231 \\ 
\midrule
 TRELLIS~\cite{xiang2025structured} & \multirow{3}{*}{2 views} & 0.686 & 13.98 & 0.346 \\
 Amodal3R~\cite{wu2025amodal3r} &  & 0.819 & 16.52 & 0.238 \\ 
  Ours & & 0.837 & 16.99 & 0.212  \\ 
\midrule
 TRELLIS~\cite{xiang2025structured} & \multirow{3}{*}{4 views} & 0.674 & 13.66 & 0.357 \\
 Amodal3R~\cite{wu2025amodal3r} &  & 0.832 & 16.64 & 0.236 \\ 
  Ours & & 0.838 & 16.98 & 0.208  \\ 
\bottomrule
\end{tabular}
\label{tab: reb_rec}
\vspace{-.2in}
\end{table}
}

\subsection{Experimental Settings}
\noindent\textbf{Dataset.} We use 3D-FUTURE~\cite{fu20213d} and Amazon Berkeley Objects (ABO)~\cite{collins2022abo} for 3D consistent occluded object simulation. 3D-FUTURE contains 9472 objects, and ABO contains 4485. We train our network with all objects combined. For each object, we render 150 views while filtering out camera elevation angles lower than -75 degrees or higher than 75 degrees, which results in around 90 views per object that are close to real-world scenarios and captures.
We select views whose occluded regions take up from 20\% to 60\% during training.
For evaluation, we use the Google Scanned Object (GSO)~\cite{downs2022google} dataset for object-level test.
In addition, we sample from Hypersim~\cite{roberts2021hypersim} dataset to simulate \emph{in-the-scene} scenraios; COCO~\cite{lin2015microsoftcococommonobjects} and Mip-NeRF 360~\cite{barron2022mipnerf360} dataset to evaluate our \emph{in-the-wild} performance in Sec.~\ref{subsec:inthewild}. 

\noindent\textbf{Implementation Details.}
We make the entire sparse structure generation stage (stage 1) of the model trainable and use 8 RTX 6000 GPUs in training. We use $\pi^{3}$~\cite{wang2025pi} as the multi-view stereo model and normalize the coordinates of partial point clouds. Then we perform voxelization with a grid size of $64^3$. We adopt the AdamW~\cite{loshchilov2017decoupled} optimizer with a learning rate of 5e-5, and Classifier-Free-Guidance~\cite{ho2022classifier} with a drop rate of 0.1 during training. We train the model for 12 epochs with a global batch size of 32, which takes approximately 16 hours to complete. Please refer to the supplementary for more details in the training, inference, and data engine settings.


\subsection{Main Results}
\noindent\textbf{Comparison Methods} We compare our method with various 3D object generation methods equipped with the latest 2D completion methods. 
First, to validate the 3D amodal completion ability with unposed sparse views of \name{}, we employ TRELLIS~\cite{xiang2025structured}, Amodal3R~\cite{wu2025amodal3r}, FreeSplatter~\cite{xu2025freesplatter}, and the most recent SAM3D~\cite{chen2025sam} (implemented MV-SAM3D for sparse-view scenarios) for comparison. We utilize \cite{ao2025open} to perform 2D amodal completion at sparse views.
Second, to demonstrate that \name{} is also effective and can generate high-quality 3D assets in real-world scenarios, we show \emph{in-the-wild} and \emph{in-the-scene} qualitative results, compared with Amodal3R~\cite{wu2025amodal3r} and SceneGen~\cite{meng2025scenegen}, respectively.

\begin{figure}[t]
    \centering
\includegraphics[width=0.8\linewidth]{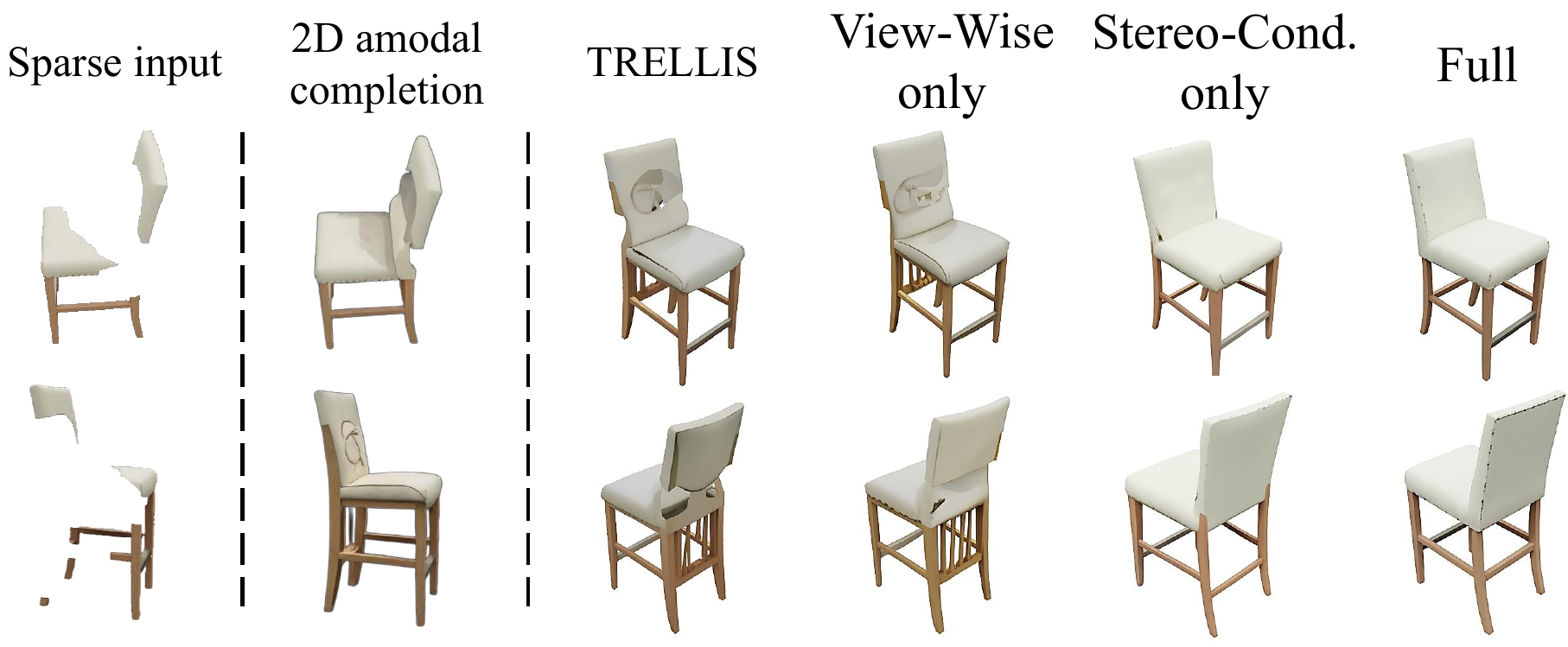}\\
    \vspace{-.05in}
    \caption{Visual comparisons from the ablation studies showing the impact of each proposed model design.}
    \vspace{-.15in}
    \label{fig:ablation_structure}
\end{figure}

\noindent\textbf{Quantitative Results.}
We present quantitative results tested on the GSO dataset. We first render 3D consistent masks and then follow the inference proxy (Sec.~\ref{method:training}) to generate all 1030 objects in the dataset. For a fair comparison, we select those views whose visible parts occupy from 30\% to 70\% of the whole rendering.
To validate the effectiveness and generalizability of our \name{}, we compare our method with others on different numbers of sparse view settings, \ie, 1, 2, and 4 views. 
\junwei{We note that SAM3D is evaluated solely under the single-view setting, consistent with its original design configuration and the official implementation.}
As shown in Tab.~\ref{tab: exp_metric}, we employ Fr\'echet Inception Distance (FID)~\cite{heusel2017gans}, Kernel Inception Distance (KID)~\cite{binkowski2018demystifying} to evaluate the quality of rendered images; CLIP-Score~\cite{radford2021learning} to measure the feature similarity; Minimum Matching Distance (MMD) and Coverage (COV)~\cite{nichol2022point} to evaluate the geometry quality.
\junwei{Moreover, we evaluate the Perceptual Coherence Score (denoted as PCS) of the generated 3D objects using a VLM~\cite{bai2025qwen3}. See supplementary material for detailed explanations of applied metrics.}

\begin{table}[t]
    \small
    \centering
    \caption{Ablation study on different proposed modules. Red indicators represent performance \textbf{drops} compared to the full model.}
    \vspace{-.05in}
    \renewcommand\arraystretch{1}
    \begin{tabular}{l|lll}
    \toprule
        Model Structure & FID $\downarrow$  & MMD{(\scriptsize \textperthousand)} $\downarrow$ & COV{\scriptsize(\%)}$\uparrow$ \\
    \midrule
        Ours full & 32.12  &  5.50 & 39.08 \\
    \midrule
        - Gating MLP & 32.87{\scriptsize\color{red}+0.55} & 5.58{\scriptsize\color{red}+0.08} & 37.68 {\scriptsize\color{red}-1.40} \\
        - View-Wise CA & 38.64{\scriptsize\color{red}+6.42} &  5.76{\scriptsize\color{red}+0.26} & 38.34 {\scriptsize\color{red}-0.74} \\
        - Stereo-Cond. CA & 42.92{\scriptsize\color{red}+10.80} & 5.91{\scriptsize\color{red}+0.41} & 35.26 {\scriptsize\color{red}-3.82} \\
        - all proposed & 46.23{\scriptsize\color{red}+14.11} & 5.94{\scriptsize\color{red}+0.44} & 34.42 {\scriptsize\color{red}-4.66} \\
    \bottomrule
     \end{tabular}
    \label{tab:ablation_structure}
    \vspace{-.1in}
\end{table}

\begin{table}[t]
    \small
    \centering
    \caption{Ablation study on \name{}'s ability with number of input views, we test up to 20 views. The red/green indicator represents performance gain/loss, respectively.}
    \vspace{-.05in}
    \renewcommand\arraystretch{1}
    \begin{tabular}{c|lll}
    \toprule
        Num. of Views & FID $\downarrow$ & CLIP-Score{\scriptsize(\%)} $\uparrow$& MMD{\scriptsize(\textperthousand)} $\downarrow$  \\
    \midrule
        1 & 33.91 & 82.23  & 5.52  \\
        2 & 32.12{\scriptsize\color{red}-1.79} & 82.33{\scriptsize \color{red} +0.10}  & 5.50{\scriptsize \color{red} -0.02}  \\
        4 & 30.73{\scriptsize\color{red}-1.39}& 82.53{\scriptsize \color{red} +0.20}  & 5.48{\scriptsize \color{red} -0.02}  \\
        10 & 29.56{\scriptsize \color{red} -1.17} & 82.98{\scriptsize \color{red} +0.45}  & 5.45{\scriptsize \color{red} -0.03} \\
        20 & 29.43{\scriptsize \color{red} -0.13} & 82.95{\scriptsize \color{green} -0.03} & 5.45{\scriptsize \color{gray} 0.00} \\
    \bottomrule
     \end{tabular}
    \label{tab:ablation_viewnum}
    \vspace{-.25in}
\end{table}

\noindent\textbf{Qualitative Results.}
We present qualitative results on the GSO dataset with simulated occlusions for amodal 3D object generation. As shown in Fig.~\ref{fig:qual_gso}, our model outperforms other methods in rendering quality and geometry consistency. Moreover, by bridging 2D priors and 3D coherence, \name{} shows better potential in uncommon and complex amodal object completion.
We showcase more visualization results in Fig.~\ref{fig:comp_coarse}, together with the 2D amodal completion results to demonstrate \name{}'s ability in fixing image-level structural inconsistencies in the 3D space.

\noindent\textbf{Generation Faithfulness.}
We also compute the average of visibility-masked SSIM~\cite{wang2004image}, LPIPS~\cite{zhang2018unreasonable}, and PSNR over observable regions \junwei{at their corresponding views}, focusing on \junwei{the generation} fidelity to the \junwei{visible parts of} sparse inputs in amodal scenarios. As shown in Tab.~\ref{tab: reb_rec}, compared to previous SoTA methods and baselines, our method achieve superior performance in terms of \junwei{preserving observed visual details within the generated 3D objects}.

\vspace{-.15in}

\subsection{Ablation Study}
\noindent\textbf{Structure Design.}
We conduct ablation studies to assess the effectiveness of different module designs.
Tab.~\ref{tab:ablation_structure} reports quantitative results, and Fig.~\ref{fig:ablation_structure} shows visual comparisons.
With \emph{View-Wise Cross Attention}, our method performs notably better under sparse views, demonstrating better rendering quality with more reasonable sparse structure as prior for texture generation.
Moreover, incorporating \emph{Stereo-Conditioned Cross Attention} for explicit geometry guidance yields large benefits in the geometric metrics (MMD).
As shown in Fig.~\ref{fig:ablation_structure}, MVS conditioning helps \name{} produce more faithful and structurally consistent geometry, even when the 2D completions are inconsistent or contain artifacts.

\noindent\textbf{Number of Views.}
We investigate how the number of input views influence our method. As shown in Tab.~\ref{tab:ablation_viewnum}, the provided metrics remain stable or exhibit slight improvements, indicating that our method can handle semi-dense settings with more input images, which demonstrates the adaptability of our work. 

\noindent\textbf{2D Amodal Completion.} We conduct experiments on how our model performs with different 2D amodal completion methods. As shown in Tab.~\ref{tab:ablation_2damodal}, we can see that for different methods, our \name{} is able to produce consistent and robust results that are comparable in numerical metrics. Additionally, it is intuitive that stronger completion methods produce better results, which shows the potential and adaptability of our \name{} with more advanced techniques in the future.

\noindent\textbf{Errors in MVS.}
In recent MVS models, the reconstructed point clouds occasionally exhibit misaligned or failing structures. Therefore, we analyze how such MVS errors affect our generation pipeline. As shown in Fig.~\ref{fig:ablation_mvs}, when the reconstructed stereo is geometrically implausible, our model can be less influenced by the stereo conditioning and perform self-correction with the 3D generative prior.

\begin{figure*}[t!]
    \centering
    \includegraphics[width=1.0\linewidth]{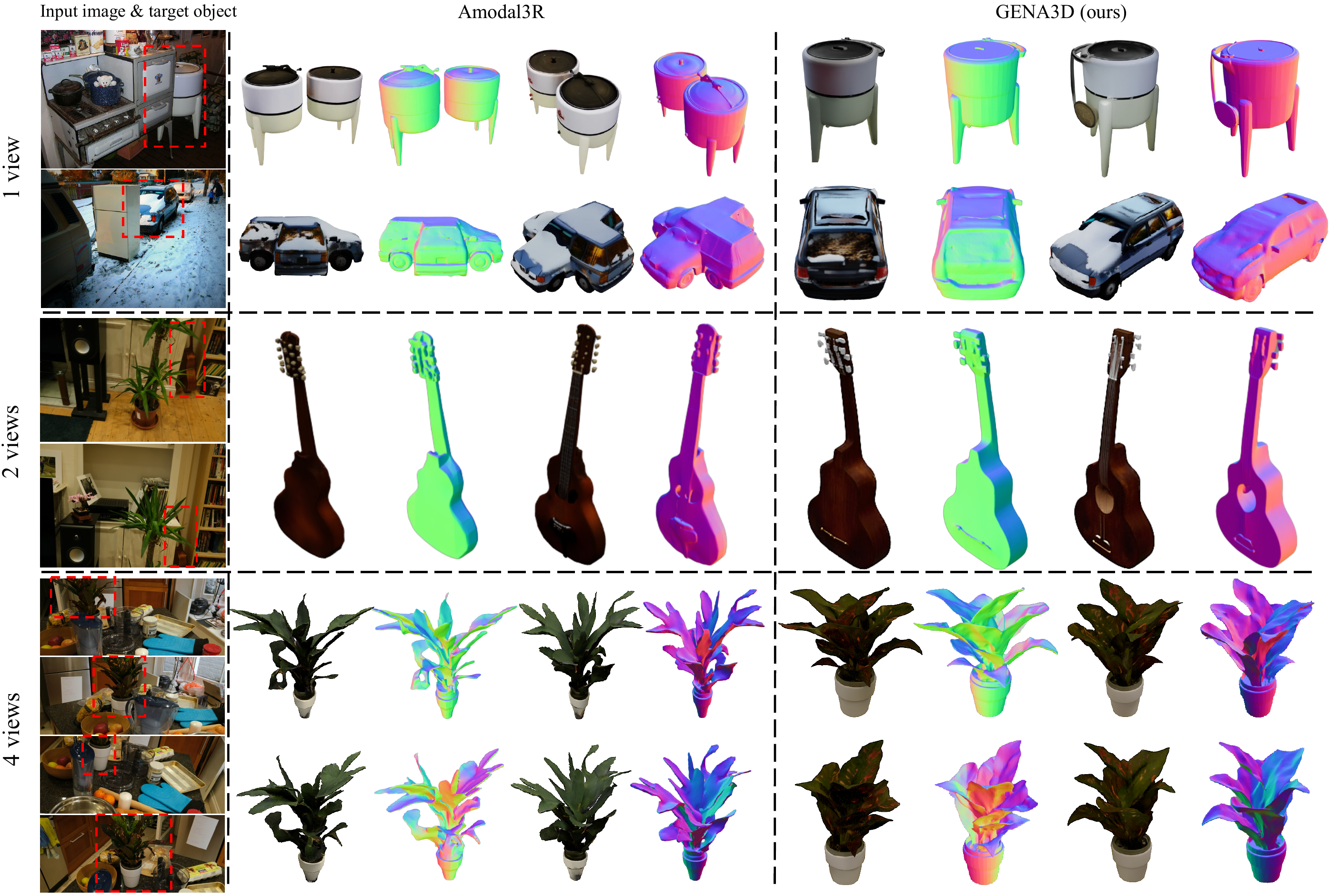}\\
    \caption{In-the-wild 3D amodal generation results. Single-view examples are from the COCO~\cite{lin2015microsoftcococommonobjects} dataset; sparse-view examples are from the Mip-NeRF 360~\cite{barron2022mipnerf360}.}
    \label{fig:inthewild}
    \vspace{-.05in}
\end{figure*}

\begin{table}[t]
    \small
    \centering
    \caption{Ablation study on different 2D Amodal Completion methods. We present results generated with 2 views.}
    \vspace{-.1in}
    \renewcommand\arraystretch{1}
    \begin{tabular}{c|ccc}
    \toprule
        2D Amodal Comp. & FID $\downarrow$ & CLIP-Score{(\scriptsize \%)} $\uparrow$& COV{(\scriptsize \%)} $\downarrow$  \\
    \midrule
        pix2gestalt~\cite{ozguroglu2024pix2gestalt} & 38.61 & 82.07  & 37.26  \\
        Flux inpainting~\cite{flux2024} & 33.07 & 84.26  & 39.49 \\
        OAAC~\cite{ao2025open} & 32.12 & 82.33  & 39.08 \\
    \bottomrule
     \end{tabular}
    \label{tab:ablation_2damodal}
    \vspace{-.2in}
\end{table}
\vspace{-.15in}
\begin{figure}[!ht]
    \centering
    \includegraphics[width=0.95\linewidth]{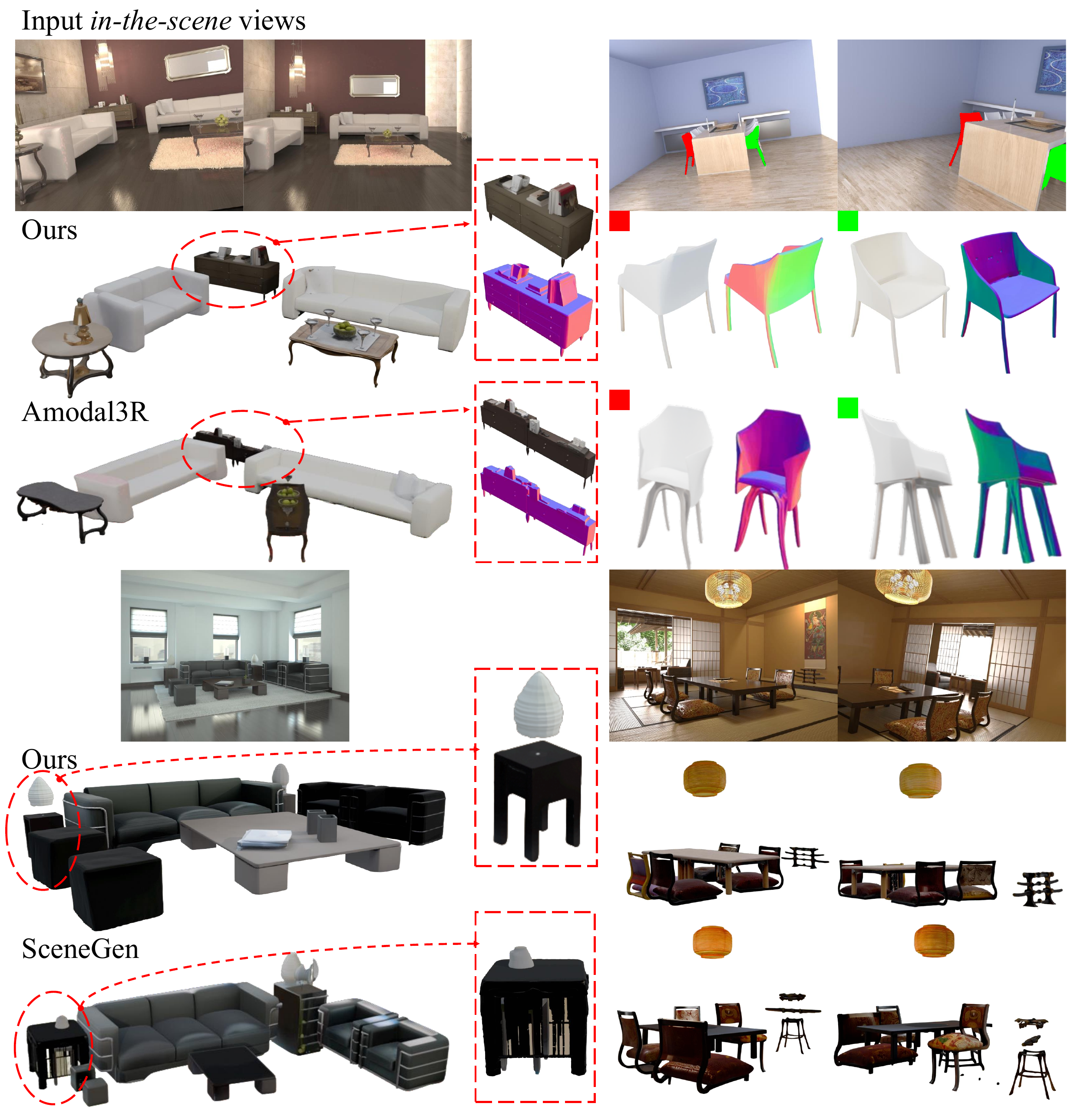}\\
    \vspace{-.1in}
    \caption{In-the-scene 3D amodal generation results. Input images are from the Hypersim~\cite{roberts2021hypersim} dataset. {\color{red}{Red}} and {\color{green}{Green}} icons indicate different target objects.}
    \label{fig:inthescene}
    \vspace{-.25in}
\end{figure}


\subsection{Generation In-the-wild \& In-the-Scene}
\vspace{-.1in}
\label{subsec:inthewild}
We further evaluate our model in both \textit{in-the-wild} and \textit{in-the-scene} settings to assess its robustness and generalization ability. 
The \textit{in-the-wild} cases correspond to real-world environments with complex interactions between objects, while the \textit{in-the-scene} setting mainly involves structured indoor layouts, such as furniture arrangements or tabletop scenes. 
As shown in Fig.~\ref{fig:inthewild}, our \name{} achieves realistic 3D amodal generation across various \textit{in-the-wild} conditions.
In Fig.~\ref{fig:inthescene}, we show \textit{in-the-scene} generation results compared with ~\cite{wu2025amodal3r} and ~\cite{meng2025scenegen}, indicating the potential of our model in object-level scene generation under the sparse-view and severe occlusion setting.
These results demonstrate that with strong 2D priors and 3D coherence, our method can generalize across different domains, retain visible partial information, and faithfully hallucinate unseen parts.

\vspace{-.15in}
\section{Conclusion}
\label{sec:conclusion}
\vspace{-.1in}
In this paper, we propose \name, a 3D object generator capable of generating amodal 3D objects from arbitrary sparse and partial occluded observations. Using only object-level occlusion training data and simulated 2D amodal completion, \name{} can seamlessly integrate with 2D amodal completion methods and multi-view stereo models, enabling generalization across both synthetic and real-world scenarios and paving the way for high-quality 3D generation in diverse applications.
While \name{} relies on priors learned from generative models to infer missing regions, which may look plausible but not always match the ground-truth geometry, the approach still represents a meaningful step forward in 3D generation, demonstrating the potential of \junwei{bridging strong 2D generative priors and 3D geometry coherence under real-world amodal scenarios}.
\newpage

%
%
\bibliographystyle{splncs04}
\bibliography{main}

\clearpage
\setcounter{page}{1}
\setcounter{section}{0}
\renewcommand\thesection{\Alph{section}}
\counterwithin{figure}{section}
\counterwithin{table}{section}
\renewcommand\thefigure{\thesection.\arabic{figure}}
\renewcommand\thetable{\thesection.\arabic{table}}

\begin{center}
    {\Large\textbf{Supplementary Material}}
\end{center}

\section{More Implementation Details}
\label{sec:more_impl}
We give more implementation details on the Network Architecture (Sec.~\ref{supp_subsec:network_archi}) and Data Engine (Sec.~\ref{supp_subsec:dataengine}) in this section. Moreover, we present a comprehensive and detailed evaluation setting for our quantitative and qualitative results.

\subsection{Network Architecture}
\label{supp_subsec:network_archi}
In this section, we specify some designs of the proposed network structure.

In the Stereo-Conditioned Cross Attention, 
we train this module from scratch and adopt the Xavier initialization~\cite{glorot2010understanding}. 
We normalize the point cloud coordinates to [-0.5, 0.5], and then use a grid size (voxel size) of $1/64$ for point cloud voxelization. Moreover, we perform certain noisy point cloud filtering during this process: for each point $p$ in the point cloud, we apply a neighborhood search within a fixed spatial radius of $1/64$, and if there are fewer than 5 points within this radius range, we simply discard the outlier point to achieve cleaner partial point clouds.

The Geometry-Guided Gating MLP is consisted of: \textbf{1) the first linear layer} that compresses the latent channels to $D / 2$, followed by a ReLU activation function; \textbf{2) another linear layer} that unifies the attention score, followed by a Sigmoid activation function.

Since the TRELLIS model uses a default 12 sampling steps for generation, for multi-view image conditions that are not divisors of 12, we increase the number of sampling steps to meet their minimum requirements. 

\subsection{Data Engine}
\label{supp_subsec:dataengine}
We provide additional details for the 3D consistent mask simulation part mentioned in Sec.~\ref{method:training}.
When diffusing 3D consistent masks, we give the range of coverage ratio as $20\% - 60\%$. We denote the final generated 3D consistent mask as $M_{3d}$.
To synthesize the occlusion masks used for training, we first render the occlusion and visibility parts of each object with the simulated $M_{3d}$, forming $m_{3docc}$ and $m_{3dvis}$. After that, we generate the bounding rectangle (box) $b_{occ}$ for the $m_{3docc}$ and take the pixels inside as the initial canvas. The final occlusion mask is generated by performing erosion $1-3$ times with a kernel size of $k=13$, and simultaneously setting $m_{3dvis}$ as all zeros to keep the visibility of simulated objects.

\subsection{Evaluation setting}
\label{supp_subsec:evaluation}
\noindent\textbf{Image-level Quality \& Similarity.}
We evaluate the image-level quality of the renderings using we employ Fr\'echet Inception Distance (FID)~\cite{heusel2017gans}, Kernel Inception Distance (KID)~\cite{binkowski2018demystifying} as mentioned in the main paper. 
Additionally, we use CLIP-Score~\cite{radford2021learning} with ViT-B/32 model to calculate the feature-level cosine similarity between the rendered images and the input images.
Specifically, for each object, we render 8 views using camera poses of a constant radius (2) and Field-of-View (40°), with azimuth angles of \{0°, 45°, 90°, 135°, 180°, 225°, 270°, 315°\} and a constant elevation angle of 15°. We calculate the score for each pair of corresponding views (generated and ground truth), and finally do a mean average for the metric score calculation.

\noindent\textbf{Geometry Quality.}
Moreover, we use Coverage (COV) and Minimum Matching Distance (MMD) with Chanmfer Distance~\cite{nichol2022point} to evaluate the geometry quality. We sample 4096 points from each ground truth and generated point cloud, which are obtained from the depth maps from different views using the farthest point sampling (FPS).

\noindent\textbf{Perceptual Coherence Score.}
Finally, we define a VLM~\cite{bai2025qwen3}-based evaluation metric to validate our method's ability in generating visually and geometrically coherent 3D objects under amodal scenarios.
We render the generated 3D objects at the same camera poses as what was done when computing the CLIP-Score. We then input the rendered views to the Qwen3-VL 32B model~\cite{bai2025qwen3} together with a systematic prompt, asking the model to rate the spatial and appearance coherence of the generated 3D object from 0 to 100. After collecting the scores for each 3D object, we do a mean average to calculate the final score. The systematic prompt is shown below:
\begin{promptbox}
\begin{lstlisting}[breaklines=true]
You are an expert 3D visual quality evaluator. You will be given multiple images rendered from different viewpoints of a single 3D object.

Your task is to evaluate two aspects:

1. Multi-View Coherence: Assess whether the object maintains consistent geometry, texture, color, and identity across all provided views. Penalize inconsistencies such as shape distortion, texture flickering, missing parts, or identity drift between viewpoints.

2. Appearance Quality: Assess the visual fidelity of the rendered object, including surface detail, texture sharpness, and absence of artifacts (e.g., blurriness, noise, or floaters).

Scoring guidelines:
  - 0-20:   Severe inconsistencies or artifacts, nearly unusable quality
  - 21-40:  Noticeable issues that significantly degrade the result
  - 41-60:  Moderate quality with some visible flaws
  - 61-80:  Good quality with minor issues
  - 81-100: High quality, coherent across views, visually convincing

Output format (strictly follow this):
{
    "score": <int>
}

Do not include any text outside the output format.
\end{lstlisting}
\end{promptbox}

\section{Additional Results}
\label{sec:add_results}
In this section, we present more qualitative results on the 3D amodal generation results at the single-object level in Sec.~\ref{supp_subsec:add_obj} and an occlusion degree analysis in Sec.~\ref{supp_subsec:occ} (additional results on the GSO dataset); we also give more \textit{in-the-wild} and \textit{in-the-scene} results in Sec.~\ref{supp_subsec:inthewildinthescene}.

For the efficiency evaluation, our \name{} can typically generate a single 3D object within 15 seconds, which causes a slight computation overhead compared to TRELLIS~\cite{xiang2025structured} and Amodal3R~\cite{wu2025amodal3r}. However, our method leads to much better generation results, which is worth the increased inference time.

\subsection{Occlusion Degree}
\label{supp_subsec:occ}
We randomly select 500 objects from the GSO dataset and render 3D-consistent masks at occlusion levels of 20\%, 50\%, and 75\%, as suggested by Reviewer ruEG.  
We evaluate FID, KID, and CLIP-Score using one and two input views (Tab.~\ref{tab: reb_occ}), demonstrating that our model remains robust under varying degrees of occlusion.

{\begin{table}[ht!]
\vspace{-.2in}
\centering
\small
\caption{The performance of our GENA3D under different occlusion degrees. We test under the setting of single view and 2 views, with occlusion degrees of 20\%, 50\%, and 70\%.}
\begin{tabular}{c|c|ccc}
\toprule
Occ. Degree & View Num. & FID $\downarrow$  & KID{\scriptsize(\%)} $\downarrow$  &CLIP{\scriptsize(\%)} $\uparrow$ \\
\midrule
 20\% &\multirow{3}{*}{1 view}  & 32.28 & 0.44 & 82.40  \\ 
 50\% &  & 34.16 & 0.46 & 82.08  \\ 
 70\% &  & 37.22 & 0.47 & 81.44 \\ 
\midrule
20\% &\multirow{3}{*}{2 views} & 31.98 & 0.44 & 82.54\\ 
50\% & & 33.42 & 0.46 & 82.32  \\ 
70\% & & 37.16 & 0.47 & 81.66 \\ 
\bottomrule
\end{tabular}
\label{tab: reb_occ}
\end{table}
}
\vspace{-.2in}

\subsection{Additional Single-Object Results}
\label{supp_subsec:add_obj}
\begin{figure*}[!t]
  \centering
  \includegraphics[width=1.0\linewidth]{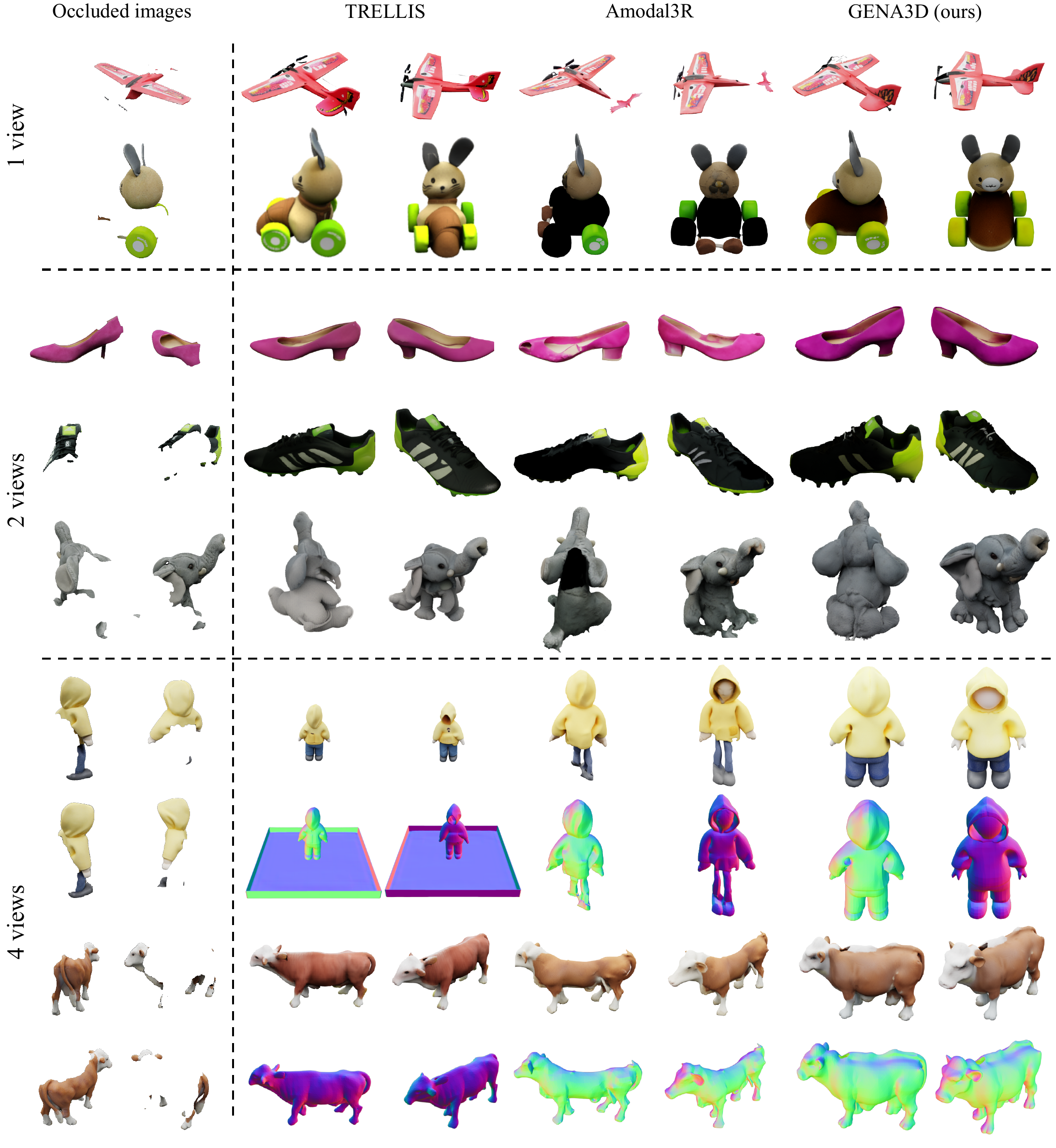}
   \caption{More qualitative results with different sparse-view settings (1, 2, and 4 views). We conduct comparisons with TRELLIS and Amodal3R on the Google Scanned Object dataset, demonstrating the effectiveness of our method.}
   \label{fig_supp:gso_comp}
\end{figure*}

\begin{figure*}[!t]
  \centering
  \includegraphics[width=0.9\linewidth]{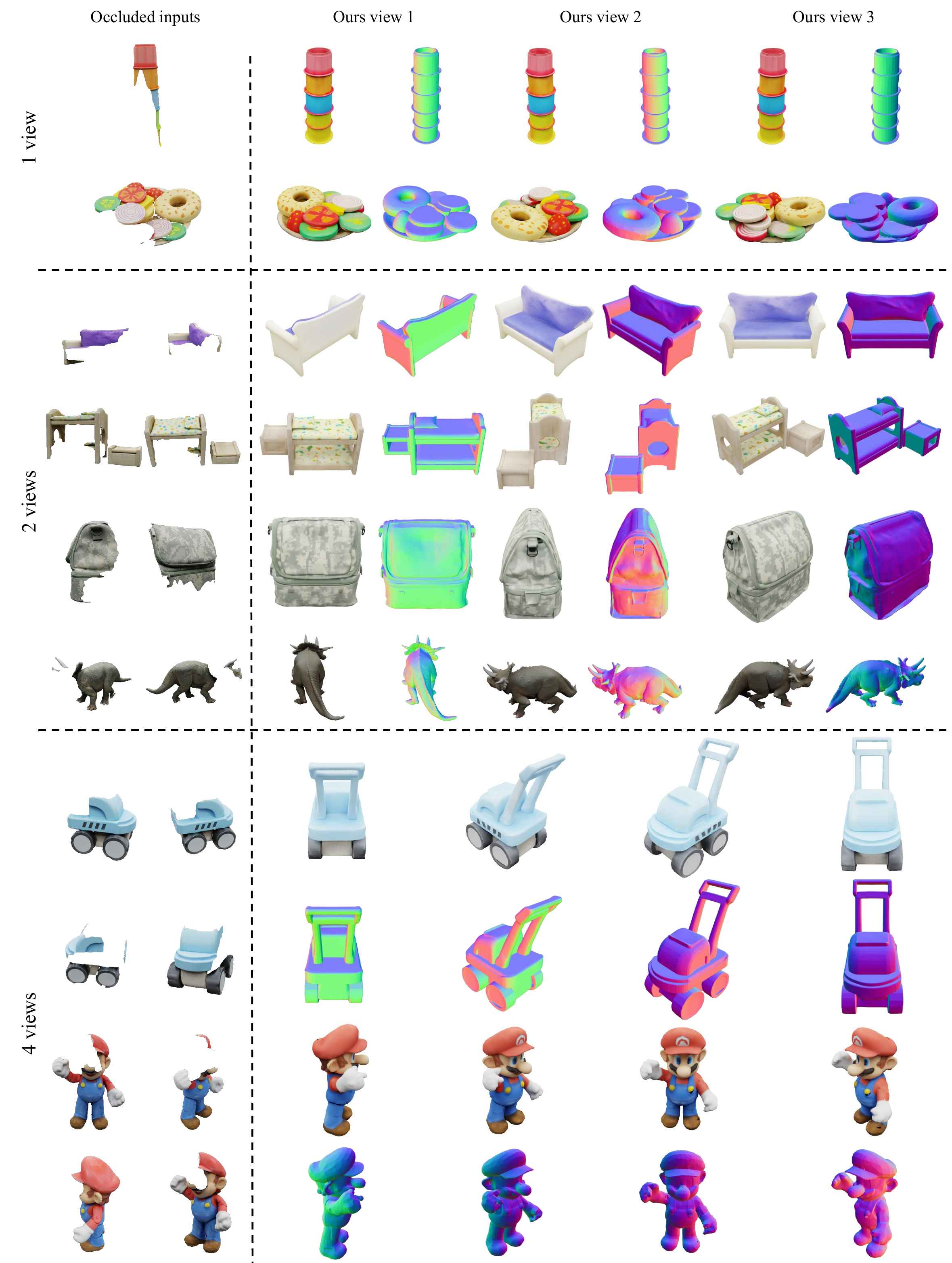}
   \caption{We present more examples generated by \name{} on the GSO dataset.} 
   \label{fig_supp:gso_more}
\end{figure*}

Fig.~\ref{fig_supp:gso_comp} and Fig.~\ref{fig_supp:gso_more} show more examples on the single-object level generation results. We present results with 1, 2, and 4 views, validating the effectiveness of our method. We still compare with TRELLIS~\cite{xiang2025structured} and Amodal3R~\cite{xu2024amodal} in this part to show that our method can fix inconsistencies and infer high-quality unobserved structures.

\subsection{Additional In-the-wild \& In-the-scene Results}
\label{supp_subsec:inthewildinthescene}
\begin{figure*}[!t]
  \centering
  \includegraphics[width=0.95\linewidth]{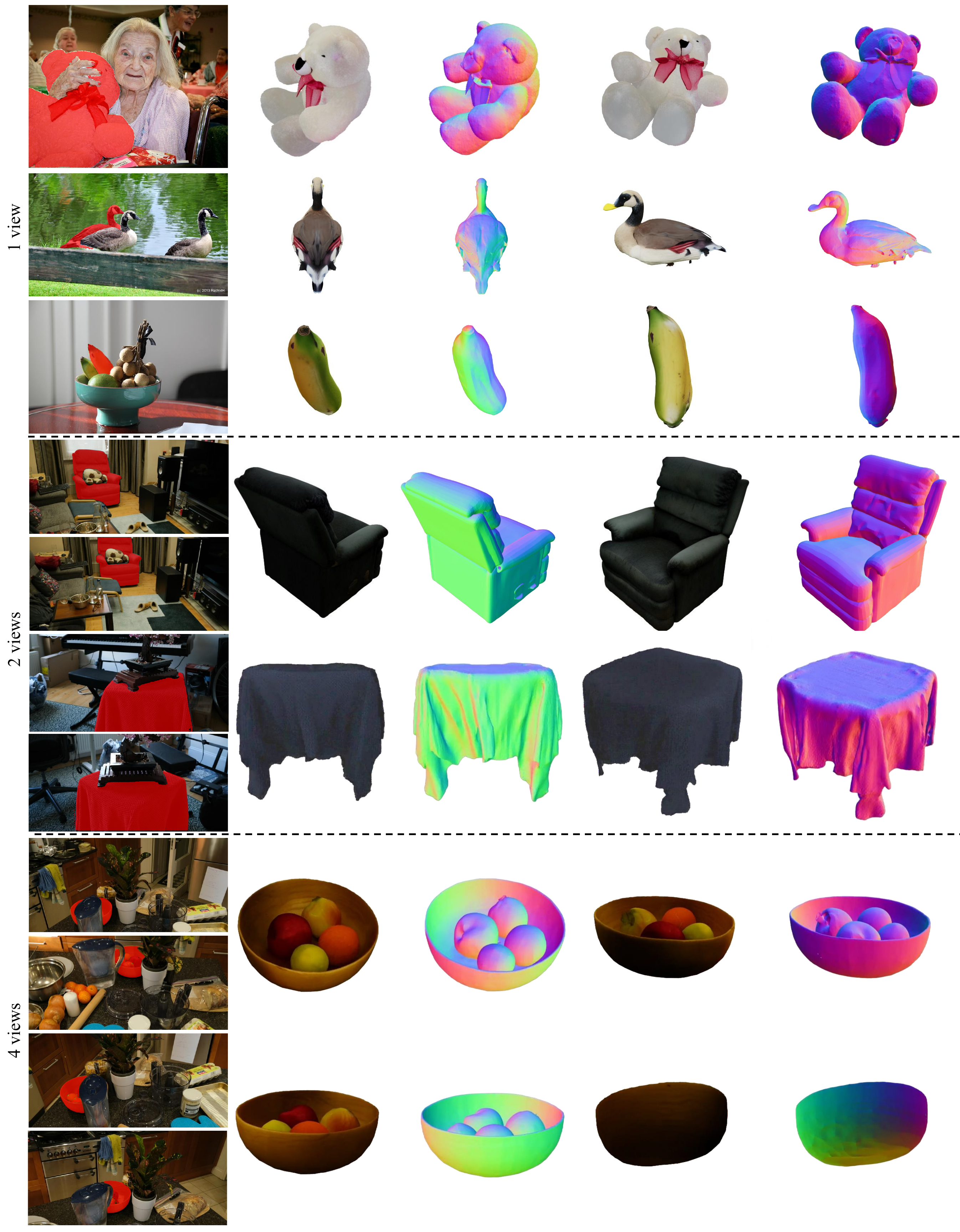}
   \caption{Additional \emph{in-the-wild} amodal 3D object generation results on the MipNeRF 360~\cite{barron2022mipnerf360} dataset.} 
   \label{fig_supp:inthewild_more}
\end{figure*}

\begin{figure*}[!t]
  \centering
  \includegraphics[width=0.9\linewidth]{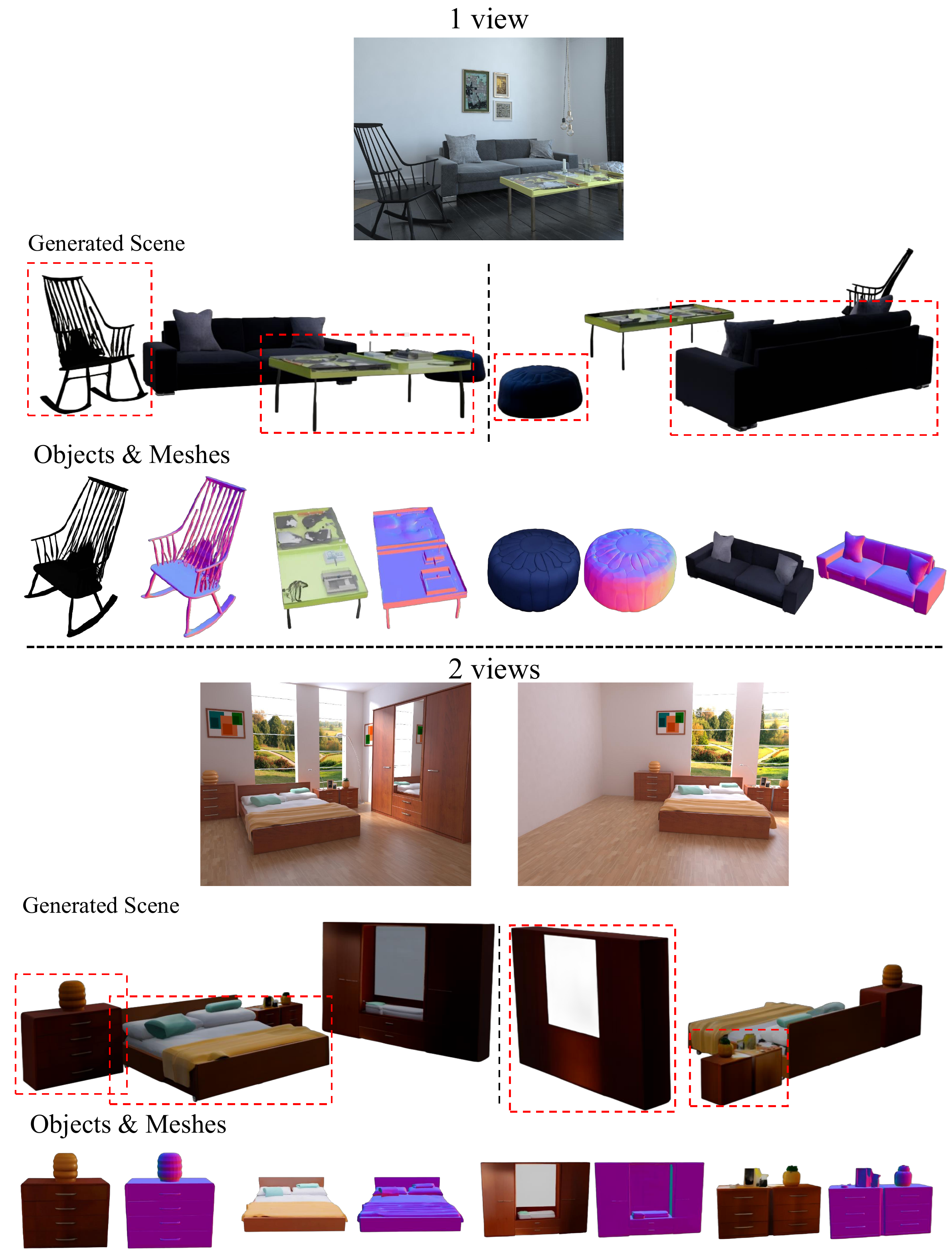}
   \caption{Additional \emph{in-the-scene} amodal 3D generation results on the Hypersim~\cite{roberts2021hypersim} dataset.} 
   \label{fig_supp:inthescene_more}
\end{figure*}

We give more \textit{in-the-wild} and \textit{in-the-scene} examples in Fig.~\ref{fig_supp:inthewild_more} and Fig.~\ref{fig_supp:inthescene_more}, showcasing that \name{} achieves superior fidelity and completeness under various occlusion conditions in real-world scenarios.

\vspace{-.1in}
\subsection{Additional comparison with SAM3D}
\label{supp_subsec:add_sam3d}
\begin{figure*}[!t]
  \centering
  \includegraphics[width=0.95\linewidth]{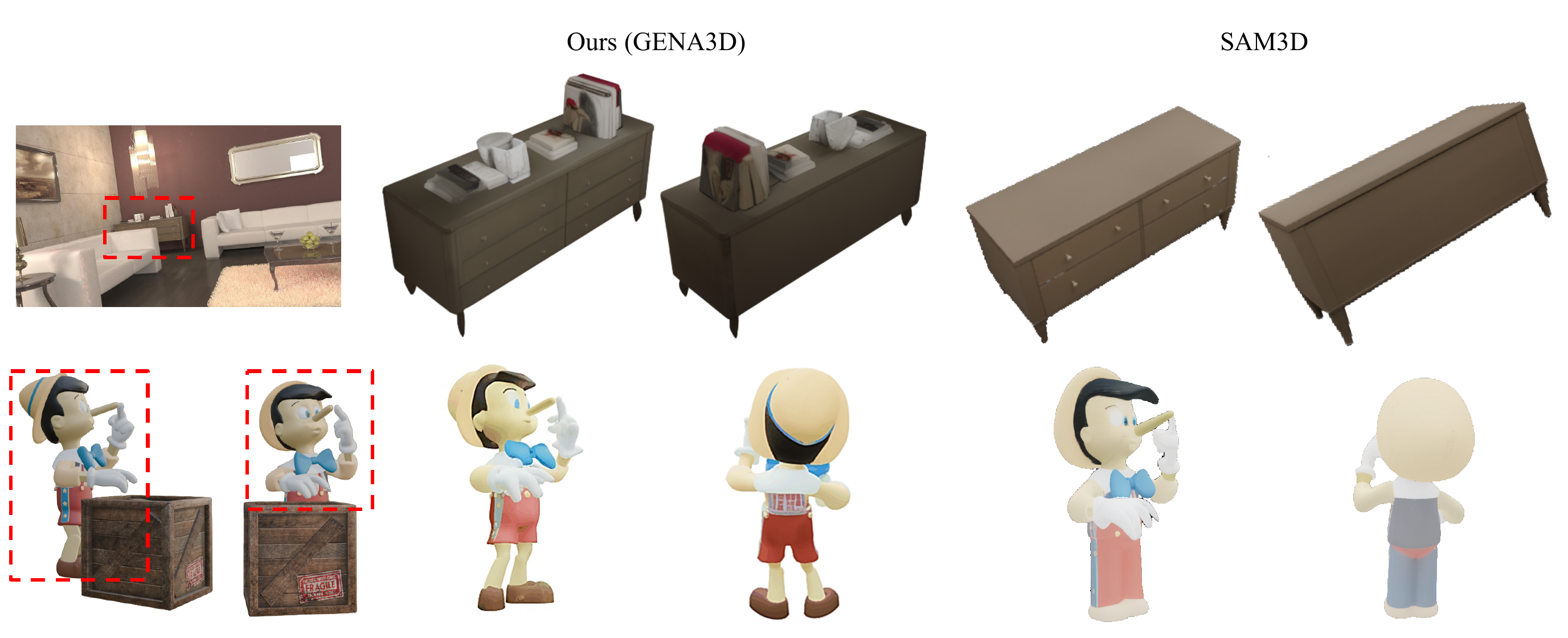}
   \caption{We present more comparisons with SAM3D.} 
   \label{fig_supp:sam3dcomp}
\end{figure*}

Additionally, we provide more comparisons between our \name{} and SAM3D in Fig.~\ref{fig_supp:sam3dcomp}. As shown in the figure, SAM3D does not generate 3D objects that are fully respective to the input observations. And even with massive training data, the generative ability is still not plausible enough. These comparisons demonstrate the importance of our design of bridging 2D priors and 3D coherence.

\subsection{Additional Ablation Studies}
We equip Amodal3R with the same OAAC front-end (Tab.~\ref{tab:isolation}). With matched 2D completion, GENA3D outperforms Amodal3R by 4.81 FID and 1.70 COV, showing the contribution of VWCA/SCCA. 
Notably, simply adding OAAC to Amodal3R degrades performance (+1.51 FID, +0.24 MMD, -1.12 COV), confirming that 2D completion alone is insufficient under this setting.
\begin{table}[h]
\vspace{-.2in}
\centering
\small
\setlength{\tabcolsep}{4pt}
\renewcommand\arraystretch{0.9} 
\caption{Results with the same 2D front-end, all results evaluated with the sampled 100 objects (same for other tables).}
\vspace{-.1in}
\label{tab:isolation}
\begin{tabular}{lccc}
\toprule
 Method & FID$\downarrow$ & MMD$\downarrow$ & COV(\%)$\uparrow$ \\
\midrule
Amodal3R                  & 35.34 & 5.52 & 38.58 \\
Amodal3R + OAAC           & 36.85{\scriptsize\color{red}+1.51} & 5.76{\scriptsize\color{red}+0.24} & 37.46{\scriptsize\color{red}-1.12} \\
\textbf{GENA3D (ours)}    & \textbf{32.04} & \textbf{5.50} & \textbf{39.16} \\
\bottomrule
\end{tabular}
\vspace{-.3in}
\end{table}

\begin{table}[h]
\centering
\small
\vspace{-.2in}
\caption{Error propagation analysis.}
\vspace{-.1in}
\label{tab:err}
\setlength{\tabcolsep}{5pt}
\renewcommand\arraystretch{0.9}
\begin{tabular}{lccc}
\toprule
 Perturbation & FID$\downarrow$ & MMD$\downarrow$ & COV(\%)$\uparrow$ \\
\midrule
2d comp. 5\%  & 31.98{\scriptsize\color{blue}-0.06} & 5.52{\scriptsize\color{red}+0.02} & 38.74{\scriptsize\color{red}-0.42} \\
2d comp. 10\% & 32.42{\scriptsize\color{red}+0.38} & 5.52{\scriptsize\color{red}+0.02} & 38.86{\scriptsize\color{red}-0.30} \\
MVS 5\% & 32.18{\scriptsize\color{red}+0.14} & 5.54{\scriptsize\color{red}+0.04} & 38.48{\scriptsize\color{red}-0.68} \\
MVS 10\% & 34.02{\scriptsize\color{red}+1.98} & 5.59{\scriptsize\color{red}+0.09} & 37.80{\scriptsize\color{red}-1.36} \\
\bottomrule
\end{tabular}
\vspace{-.3in}
\end{table}

\vspace{0.2em}
Moreover, we controllably degrade each upstream module in Tab~\ref{tab:err}:
\emph{(a) 2D completion perturb}: adding percentage gaussian noise on completed regions;
\emph{(b) MVS}: injecting Gaussian noise (percentage of the point cloud's max bounding box extent) in the depths of point clouds.
We note that GENA3D is robust to upstream noise and has acceptable degradation (+$\le$2 FID) even with 10\% perturbation.

\noindent \emph{Pose invariance:} on 100 GSO objects $\times$ 10 viewpoints, \textbf{$97.6\%$} of the generated output are in canonical axis-aligned space (measured by pairwise rotation alignment across the generations per object), validating VWCA's design rationale.
\emph{Fusion variants}: We implement uniform avg, max fusion, token-level learned attention, in Tab.~\ref{tab:fusion}, with the {\color{blue} gain} or {\color{red} loss} (same for other tables) compared to visibility-aware fusion.

\begin{table}[h]
\centering
\small
\vspace{-.2in}
\caption{Ablation study on different fusion strategies.}
\label{tab:fusion}
\setlength{\tabcolsep}{5pt}
\renewcommand\arraystretch{0.9}
\begin{tabular}{lccc}
\toprule
Fusion method & FID$\downarrow$ & MMD$\downarrow$ & COV(\%)$\uparrow$ \\
\midrule
uniform avg.  & 32.22{\scriptsize\color{red}+0.18} & 5.54{\scriptsize\color{red}+0.04} & 38.42{\scriptsize\color{red}-0.74} \\
max fusion & 34.66{\scriptsize\color{red}+2.62} & 5.68{\scriptsize\color{red}+0.18} & 36.94{\scriptsize\color{red}-2.22} \\
token-level learned attn. & 34.58{\scriptsize\color{red}+2.54} & 5.76{\scriptsize\color{red}+0.26} & 37.88{\scriptsize\color{red}-1.28} \\
\bottomrule
\end{tabular}
\vspace{-.4in}
\end{table}

\subsection{Human Evaluation}
\vspace{0.2em}
We validate PCS with a human preference evaluation (Tab.~\ref{tab:hpe}). We collect direct comparisons at $K\!\in\!\{1,2,4\}$ views (30 objects $\times$ 10 raters, side-by-side GENA3D vs.\ Amodal3R). The human evaluation results match PCS at all view counts: GENA3D is preferred by large margins, especially as the number of views increases, with a pairwise inter-rater agreement of 77.8\%. This validates GENA3D's advantage in human perceptual evaluation. 
\begin{table}[h]
\centering
\small
\vspace{-.2in}
\caption{Human preference evaluation.}
\label{tab:hpe}
\setlength{\tabcolsep}{5pt}
\renewcommand\arraystretch{0.9}
\begin{tabular}{lccc}
\toprule
 Win Rate (\%) & $K\!=\!1$ & $K\!=\!2$ & $K\!=\!4$ \\
\midrule
GENA3D    & \textbf{65.4} & \textbf{74.1} & \textbf{78.2} \\
Amodal3R  & 34.6 & 25.9 & 21.8 \\
\bottomrule
\end{tabular}
\vspace{-.2in}
\end{table}

\vspace{-.2in}

\FloatBarrier

\section{Failure Modes}
\begin{figure}[ht!]
    \vspace{-0.1in}
    \centering
    \includegraphics[width=\linewidth]{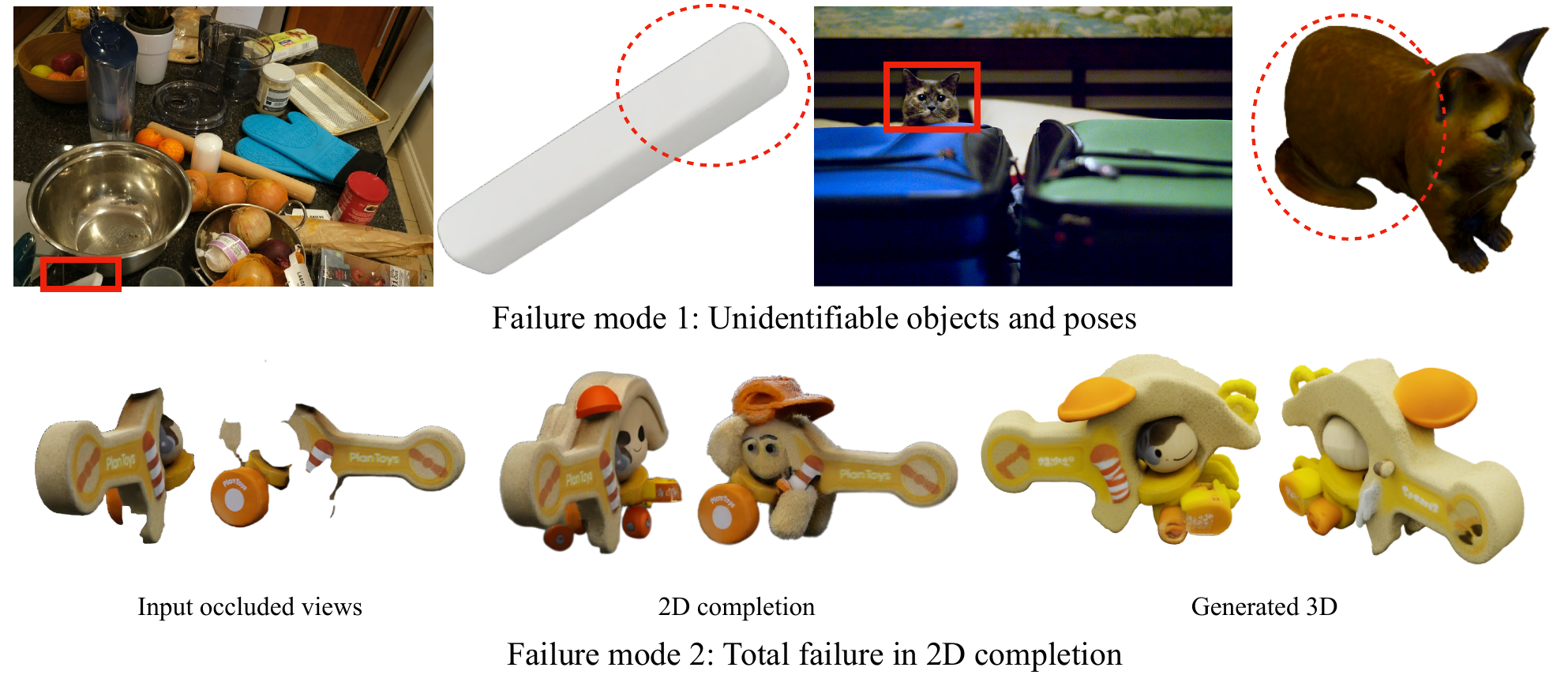}\\
    \caption{We demonstrate two main failure modes.}
    \label{fig:failure}
    \vspace{-0.1in}
\end{figure}

Fig.~\ref{fig:failure} show two major failure modes: (i) objects with unidentifiable appearance or poses, often caused by extreme ($>$85\%) occlusions or hidden discriminative parts of objects; (ii) the 2D completion module produces severely degraded inputs; our GENA3D tolerates moderate noise but cannot recover when 2D completion catastrophically fails. 
Future work includes joint training with the cross-view 2D completion stage to mitigate cascading failures.

\end{document}